\documentclass[twoside,11pt]{article}
\usepackage[round,authoryear]{natbib}
\usepackage[T1]{fontenc}
\usepackage[utf8]{inputenc}
\usepackage[margin=0.93in]{geometry}
\usepackage{amsmath,amssymb,amsthm,mathtools,mathrsfs}
\usepackage{palatino}
\usepackage{xcolor}
\usepackage{graphicx}
\usepackage[ruled,vlined]{algorithm2e}
\usepackage{microtype}
\usepackage{enumitem}
\usepackage{hyperref}
\hypersetup{colorlinks=true,linkcolor=blue,citecolor=blue,urlcolor=blue}
\newcommand{\E}{\mathbb{E}}

\renewcommand{\P}{\mathbb{P}}

\newcommand{\R}{\mathbb{R}}
\newcommand{\cA}{\mathcal{A}}
\newcommand{\cB}{\mathcal{B}}
\newcommand{\cF}{\mathcal{F}}
\newcommand{\cH}{\mathcal{H}}
\newcommand{\cS}{\mathcal{S}}

\DeclareMathOperator*{\argmin}{arg\,min}
\DeclareMathOperator*{\argmax}{arg\,max}
\DeclareMathOperator{\Var}{Var}
\DeclareMathOperator{\Cov}{Cov}
\newtheorem{theorem}{Theorem}[section]
\newtheorem{lemma}[theorem]{Lemma}
\newtheorem{proposition}[theorem]{Proposition}
\newtheorem{corollary}[theorem]{Corollary}
\theoremstyle{definition}
\newtheorem{assumption}{Assumption}
\newtheorem{remark}{Remark}

\title{Online V-Learning with Adaptive Longitudinal Data}
\author{
  Aniruddhan Ganesaraman\\
  Department of Statistics and Operations Research\\
  University of North Carolina at Chapel Hill\\
  \texttt{aniruddhan\_ganesaraman@unc.edu}
  \and
  Michael R. Kosorok\\
  Department of Biostatistics\\
  University of North Carolina at Chapel Hill\\
  \texttt{kosorok@bios.unc.edu}
}

\date{}
\begin{document}
\maketitle
\begin{abstract}
Reinforcement learning methods for estimating optimal dynamic treatment regimes are increasingly applied in mobile health (mHealth), where treatment decisions are made continually. \citet{Luckett2020} proposed V-learning, which estimates an optimal policy from offline trajectories. In practice, mHealth policies are updated online. Batch updating, in which the policy is re-estimated after each block of new data, is a tractable middle ground between offline learning and updating after every observation. However, because each batch's treatment policy is estimated from data pooled across patients in earlier batches, the resulting data are adaptively collected and patient trajectories are dependent, a setting that offline theory does not address. We develop statistical theory for V-learning in this setting with varying, deterministic batch sizes. For each fixed batch, as the number of patients grows, we establish uniform consistency of the estimated value function, consistency of the estimated policy, a functional central limit theorem over the policy class, asymptotic normality of the estimated value of the learned policy, and a consistent variance estimator. To first order, the online procedure performs as well as an oracle that deploys the optimal policy after the first batch. With the number of patients fixed and the number of batches growing, the estimated value function and policy remain consistent under a persistent identifiability condition. In simulations, as the number of patients grows with the number of batches fixed, confidence intervals attain near-nominal coverage under adaptive assignment and the online procedure closely matches the oracle. With a small cohort followed over many batches, the learned policies improve steadily.
\end{abstract}

\section{Introduction}\label{sec:introduction}

Mobile health (mHealth) interventions deliver treatment through wearable devices and smartphones, often several times a day over periods of weeks or months. Because decisions are frequent and the follow-up period has no natural end, such interventions are naturally formulated as infinite-horizon sequential decision problems, and reinforcement learning methods for estimating optimal dynamic treatment regimes are increasingly used to personalize them. \citet{Luckett2020} proposed V-learning for this setting. V-learning estimates the state-value function of each candidate policy from an importance-weighted Bellman estimating equation and maximizes the estimated value over a prespecified class of randomized policies, with values defined relative to a reference distribution of clinical interest.

The theory of \citet{Luckett2020} concerns offline estimation, in which the policy is learned from trajectories collected under a fixed behavior policy. In practice, mHealth policies are updated as data accrue, and deployed algorithms commonly do so in batches \citep{Trella2025}. Batch updating, in which the policy is re-estimated after each block of new observations, is a tractable middle ground between offline learning and updating after every decision. It also changes the statistical problem in two ways. First, each batch is generated by a policy estimated from earlier batches, so the data are collected adaptively. Second, because that policy is estimated by pooling data across patients, patient trajectories are no longer independent. Offline theory addresses neither feature, and it is well documented that adaptive data collection can invalidate standard inference \citep{ZhangJansonMurphy2020,ZhangJansonMurphy2022}.

We develop statistical theory for cumulative online V-learning with deterministic, possibly varying, batch sizes, accounting for the dependence induced when policies estimated from pooled patient data are repeatedly deployed to the same cohort. We retain the estimating equation and policy-search procedure of \citet{Luckett2020}. Our analysis rests on one structural observation. When the value functions of the candidate policies admit an exact linear representation, the importance-weighted Bellman estimating function has conditional mean zero under any behavior policy. Deploying a newly estimated policy therefore changes the distribution of subsequent states, but not the population target, and the estimating functions form martingale difference sequences, without any stationarity assumption on the state process.

With the number of batches fixed and the number of patients growing, we establish consistency, a functional central limit theorem over the policy class, first-order oracle equivalence, and inference for the value of the learned policy using a consistent patient-level variance estimator. With the number of patients fixed and the number of batches growing, we show that the value-function estimators and the estimated policy remain consistent almost surely under a persistent identifiability condition, so that even a small cohort suffices when it is followed for long enough. 

The martingale structure alone does not complete the analysis; three steps require arguments beyond those available for offline data. First, the patient histories are dependent, because each patient's later actions depend on a policy estimated from all patients' earlier data, so the law of large numbers for independent trajectories that underlies the theory of \citet{Luckett2020} does not apply. We show by induction over batches that the empirical distributions of the histories nonetheless converge to a deterministic limit (Theorem~\ref{thm:consistency}), which identifies the limiting covariance. Second, the policy is selected from the data, so inference for its value requires a central limit theorem that holds uniformly over the policy class. We obtain one for the martingale scores by combining the Azuma--Hoeffding inequality with a chaining argument (Lemma~\ref{lem:equicontinuity}). Third, with the cohort fixed, information accrues only through decision times within patients, which are neither independent nor stationary. We establish almost-sure consistency, uniformly over the policy class, through a uniform strong law for bounded martingale differences (Lemma~\ref{lem:longrun-slln} and Theorem~\ref{thm:longrun-consistency}).

The paper is organized as follows. Section~\ref{sec:related-work} reviews related work. Section~\ref{sec:online} describes online V-learning, its population target, and our assumptions. Sections~\ref{sec:fixed-batch} and~\ref{sec:fixed-n} present the results for a fixed number of batches and a fixed number of patients, respectively. Section~\ref{sec:simulations} reports simulation studies, and Section~\ref{sec:discussion} concludes. Proofs are given in the appendix.

\section{Related Work}\label{sec:related-work}

\paragraph{Policy Learning over Infinite Horizons.} Estimating a value function through the Bellman equation with a linear basis is a long-standing idea in reinforcement learning, going back to least-squares temporal-difference (LSTD) methods. \citet{LagoudakisParr2003} extended it to control through least-squares policy iteration, which alternates between estimating the state-action value function of the current policy and improving that policy. A practical advantage of the approach is that a single batch of transitions can be reused to evaluate every candidate policy. For off-policy LSTD, \citet{Yu2012} establishes almost-sure convergence under general conditions on the underlying Markov process, a result we return to when the number of batches grows.

In the statistics literature, the problem has been studied primarily in the context of mobile health, where treatment decisions are made frequently and the follow-up period has no fixed end point. \citet{ErtefaieStrawderman2018} construct dynamic treatment regimes over indefinite horizons using greedy gradient Q-learning, and \citet{LiaoQiWanKlasnjaMurphy2022} study batch policy learning in average-reward Markov decision processes. The procedure we analyze is the V-learning method of \citet{Luckett2020}. V-learning offers two advantages in health applications: the policy class can be restricted to randomized policies, so that every permitted treatment is assigned with positive probability in every state, and policy values are defined relative to a reference distribution that can be chosen to reflect the patient population of clinical interest. Although \citet{Luckett2020} also proposed online updating of V-learning, their theoretical results concern offline estimation.

\paragraph{Inference after Adaptive Data Collection.}

A growing body of work shows that adaptive data collection can invalidate inferential procedures that are valid for independent data. \citet{ZhangJansonMurphy2020} show that in batched bandit experiments the usual normal approximation can fail when treatment-assignment probabilities do not concentrate. \citet{Hadad2021} construct adaptively weighted estimators that stabilize inverse-probability weights in adaptive experiments, and \citet{ZhangJansonMurphy2021} establish asymptotic normality for suitably weighted M-estimators computed from adaptively collected data. 

The work closest to ours is that of \citet{ZhangJansonMurphy2022}, who study longitudinal data collected by algorithms that pool information across individuals, which is the same source of dependence that arises in online V-learning. They show that, because the treatment policy is itself estimated from earlier data, the standard variance estimator can underestimate the true variance, and they propose an adaptive estimator with an additional term that accounts for the estimation of the policy parameters. This term vanishes under a condition they call policy invariance: the expected estimating function does not depend, to first order, on the policy parameters. 
Under exact linear specification, the Bellman estimating function has conditional mean zero under any behavior policy, so that the standard patient-level estimator remains consistent despite adaptive treatment assignment. In related work, \citet{ZhangClosserTrellaMurphy2025} connect the replicability of adaptively learned policies to the consistency and asymptotic normality of estimators computed from the resulting data, and \citet{Trella2025} describe the deployment of such an algorithm in a clinical trial. \citet{Dwivedi2025} study inference for counterfactual outcomes in sequential experiments in which treatments are assigned adaptively.

Other work considers learning, rather than only inference, from adaptively collected data. \citet{ZhanRenAtheyZhou2024} learn policies from adaptively collected contextual bandit data and prove finite-sample regret bounds for weighted augmented inverse-probability estimators. \citet{SyrgkanisZhan2026} develop moment-based estimation and inference for data collected by multistage reinforcement learning algorithms, using adaptive weights to handle time-varying conditional variances. Our setting differs from both in that the estimated policy is repeatedly deployed in a continuous Markov decision process. Instead of introducing variance-stabilizing weights, we rely on the conditional mean-zero property of the Bellman estimating function and show that the behavior policies converge at each fixed batch index.

\paragraph{Off-Policy Evaluation and Inference over Long Horizons.}

Off-policy evaluation, the estimation of a policy's value from data collected under a different policy, has received substantial attention in both reinforcement learning and statistics.  \citet{KallusUehara2020,KallusUehara2022} study efficient off-policy evaluation in Markov decision processes, including the use of marginalized density ratios and time homogeneity to address long-horizon estimation. \citet{Madhow2025} give sample-complexity results for offline policy evaluation and policy learning in tabular Markov decision processes when the policies depend adaptively on past data.

On the inferential side, \citet{Shi2022} construct confidence intervals for policy values using series approximations of the state-action value function; for data-dependent target policies, they propose a sequential value evaluation procedure that remains valid when either the number of trajectories or the number of decision points grows. \citet{WuLiWeiRinaldo2026} study temporal-difference learning with linear function approximation, deriving Berry--Esseen bounds and an online estimator of the asymptotic covariance for independent samples. \citet{LiuTuChenZhang2025} develop online estimation and inference for policy evaluation under heavy-tailed rewards and contamination, and \citet{LuLeeAoSimchiLeviLu2026} study offline inference for the optimal value. These papers treat the data-generating process as fixed and study estimation and inference for policy values. We also study the learning procedure itself, asking whether it converges when each estimated policy generates the next batch of data.

\section{Online V Learning}\label{sec:online}
\subsection{Data and estimator}
We follow $n$ patients, indexed by $i=1,\ldots,n$, without dropout. Let $S_t^i\in\cS\subset\R^p$ be the state observed immediately before treatment at time $t$, $A_t^i\in\cA=\{a_1,\ldots,a_K\}$ the assigned treatment, and $U_t^i=u(S_{t+1}^i,A_t^i,S_t^i)$ the utility. The state evolves according to an action-dependent transition kernel $P_a(s,ds')$, common across patients and time points. Write
\[
(P_af)(s)=\int \limits_{\cS}f(s')P_a(s,ds'),\qquad r_a(s)=\int \limits_{\cS}u(s',a,s)P_a(s,ds'),
\]
where $f$ is bounded and measurable. We fix a discount factor $\gamma\in(0,1)$ and define the state-value function of a stationary policy $\pi$ by
\[
V^\pi(s)=\E_\pi\!\left[\left.\sum \limits_{m=0}^\infty\gamma^m U_{t+m}\,\right|S_t=s\right].
\]
Here $\E_\pi$ refers to subsequent treatment decisions generated by $\pi$ with transition kernel $P_a$, not to the treatment probabilities that generated the observed data.

As in \citet{Luckett2020}, we consider the multinomial logistic class $\Pi=\{\pi_\beta:\beta\in\cB\}$, where $\beta=(\beta_1^\top,\ldots,\beta_{K-1}^\top)^\top\in\cB\subset\R^{p(K-1)}$ and
\[
\pi_\beta(a_j;s)=\frac{\exp(s^\top\beta_j)}{1+\sum \limits_{\ell=1}^{K-1}\exp(s^\top\beta_\ell)},\quad j<K,\qquad
\pi_\beta(a_K;s)=\frac{1}{1+\sum \limits_{\ell=1}^{K-1}\exp(s^\top\beta_\ell)}.
\]
The coefficient space $\cB$ is assumed to be compact with nonempty interior. We approximate the value of $\pi$ by $V(\pi,s;\theta^\pi)=\Phi(s)^\top\theta^\pi$, where $\Phi:\cS\to\R^q$ is a given feature map and $\theta^\pi\in\Theta\subseteq\R^q$.

The first batch is generated under a known policy $\pi^0$. Let $B_k\in\mathbb N$ denote the deterministic length of batch $k$, and put $T_0=0$, $T_k=\sum \limits_{j=1}^kB_j$, $N_k=nT_k$. At time $t$, let $j(t)$ be the batch index satisfying $T_{j(t)-1}<t\leq T_{j(t)}$. Once batch $j$ has ended, cumulative data through time $T_j$ produce an estimated policy $\hat\pi_n^j=\pi_{\hat\beta_n^j}$. Thus the probability used to assign treatment at time $t$ is
\begin{equation}\label{eq:behavior}
\mu_{n,t}(a;s)=\begin{cases}
\pi^0(a;s),&t\le B_1,\\
\pi_{\hat\beta_n^{j(t)-1}}(a;s),&t>B_1.
\end{cases}
\end{equation}
The policy $\mu_{n,t}$ is random, but it is determined by observations collected before the treatment assignments at time $t$.

For a candidate $\pi\in\Pi$ and $\theta\in\Theta$, set
\[
w_t^i(\pi)=\frac{\pi(A_t^i;S_t^i)}{\mu_{n,t}(A_t^i;S_t^i)},\qquad
e_t^i(\theta)=U_t^i+\gamma\Phi(S_{t+1}^i)^\top\theta-\Phi(S_t^i)^\top\theta.
\]
Following the importance-weighted Bellman estimating equation, define
\begin{equation}\label{eq:Lambda}
\Lambda_{n,k}(\pi,\theta)=\frac1{N_k}\sum \limits_{t=1}^{T_k}\sum \limits_{i=1}^n w_t^i(\pi)e_t^i(\theta)\Phi(S_t^i)
=\hat b_{\mathrm{cum},k}(\pi)-\hat W_{\mathrm{cum},k}(\pi)\theta,
\end{equation}
where
\begin{align*}
\hat W_{\mathrm{cum},k}(\pi)&=\frac1{N_k}\sum \limits_{t=1}^{T_k}\sum \limits_{i=1}^n
w_t^i(\pi)\Phi(S_t^i)\{\Phi(S_t^i)-\gamma\Phi(S_{t+1}^i)\}^\top,\\
\hat b_{\mathrm{cum},k}(\pi)&=\frac1{N_k}\sum \limits_{t=1}^{T_k}\sum \limits_{i=1}^n
w_t^i(\pi)U_t^i\Phi(S_t^i).
\end{align*}

Let $\Omega$ be a fixed positive definite $q\times q$ matrix, $J:\Theta\to[0,\infty)$ a penalty, and $\lambda_{n,k}\geq0$ a deterministic tuning parameter. Our estimator is
\begin{equation}\label{eq:theta-est}
\hat\theta_{\pi,n}^k=\argmin_{\theta\in\Theta}
\left\{\Lambda_{n,k}(\pi,\theta)^\top\Omega\Lambda_{n,k}(\pi,\theta)+\lambda_{n,k} J(\theta)\right\}.
\end{equation}

For $J(\theta)=\|\theta\|^2$ and $\Theta=\R^q$, this reduces to the usual ridge-regularized minimum distance estimator. For a fixed reference distribution $R$, put $\nu=\int \limits_{\cS}\Phi(s)\,dR(s)$ and define
\begin{equation}\label{eq:beta-est}
\hat V_{n,R,k}(\pi)=\nu^\top\hat\theta_{\pi,n}^k,\qquad
\hat\beta_n^k\in\argmax_{\beta\in\cB}\hat V_{n,R,k}(\pi_\beta),\qquad
\hat\pi_n^k=\pi_{\hat\beta_n^k}.
\end{equation}

In finite samples, the matrix $\hat W_{\mathrm{cum},k}(\pi_\beta)$ may be singular for some $\beta$. Because each estimated policy generates the next batch, the procedure must nevertheless specify a policy on every event. We therefore fix default values $\bar\theta\in\Theta$ and $\bar\beta\in\cB$. If $\inf_{\beta\in\cB}\sigma_{\min}\{\hat W_{\mathrm{cum},k}(\pi_\beta)\}=0$, we set $\hat\theta_{\pi_\beta,n}^k=\bar\theta$ for all $\beta\in\cB$ and $\hat\beta_n^k=\bar\beta$. Otherwise, the estimators are defined by \eqref{eq:theta-est}--\eqref{eq:beta-est}. 

\begin{algorithm}[H]
\DontPrintSemicolon
\caption{Cumulative online V-learning with deterministic batches}\label{alg:online}
\KwIn{$n$ patients; initial policy $\pi^0$; class $\{\pi_\beta:\beta\in\cB\}$; feature map $\Phi$; reference vector $\nu=\int\Phi\,dR$; positive definite matrix $\Omega$; penalty $J$; tuning sequence $\lambda_{n,k}$; discount $\gamma$; deterministic batch sizes $\{B_k\}$.}
Set $T_0=0$, $\hat\pi_n^0=\pi^0$\;
\For{$k=1,2,\ldots$}{
Set $T_k=T_{k-1}+B_k$ and $N_k=nT_k$\;
\For{$t=T_{k-1}+1,\ldots,T_k$}{
\For{$i=1,\ldots,n$}{
Sample $A_t^i$ from $\hat\pi_n^{k-1}(\cdot;S_t^i)$ and observe $U_t^i,S_{t+1}^i$\;
}
Record $\mu_{n,t}=\hat\pi_n^{k-1}$ as the behavior policy for time $t$\;
}

For each queried $\beta\in\cB$, compute $\hat\theta_{\pi_\beta,n}^k$ using \eqref{eq:theta-est}\;
Set $\hat\beta_n^k$ and $\hat\pi_n^k$ according to \eqref{eq:beta-est}\;

}
\end{algorithm}

\subsection{Population target and assumptions}
For each candidate policy $\pi$ and state $s$, let $V_R(\pi)=\int \limits_{\cS}V^\pi(s)\,dR(s)$ denote its value under the reference distribution $R$. We assume below that the value function has an exact linear representation $V^\pi(s)=\Phi(s)^\top\theta_0^\pi$. The resulting population objective and its maximizer are
\begin{equation}\label{eq:population-optimum}
M(\beta)=V_R(\pi_\beta)=\nu^\top\theta_0^{\pi_\beta},\qquad
\beta^*=\argmax_{\beta\in\cB}M(\beta),\qquad \pi^*=\pi_{\beta^*}.
\end{equation}
The target $\beta^*$ is defined independently of the number and lengths of the batches.

For each fixed batch $k$, the population limits of the estimating matrices depend on the distributions of patient states at times $t=1,\ldots,T_k$. We describe these distributions using the large-sample behavior of the policies implemented in successive batches.

Recall that the initial policy $\pi^0$ is fixed and known, whereas the policy used in batch $j\geq 2$ is estimated from the data collected through batch $j-1$. Under Assumption~\ref{ass:unique}, the population objective has a common, unique maximizer $\beta^*$, with corresponding policy $\pi^*=\pi_{\beta^*}$. As we establish in Theorem~\ref{thm:consistency}, for each fixed $j$, the estimated policy at the end of batch $j$ converges to $\pi^*$ as $n\to\infty$. Thus, in the large-sample limit, the behavior policy is $\pi^0$ in the first batch and $\pi^*$ in every subsequent batch. We write this limiting sequence of behavior policies as
\[
\mu_t^\circ=
\begin{cases}
\pi^0, & 1\leq t\leq B_1,\\
\pi^*, & t>B_1.
\end{cases}
\]

Let $\eta_1^\circ$ denote the common initial-state distribution. For any policy $\pi$, define the transition kernel
\[
P_\pi(s,D)=\sum \limits_{a\in\cA}\pi(a;s)P_a(s,D).
\]
The state distributions under the limiting behavior-policy sequence satisfy
\begin{equation}\label{eq:state-recursion}
\eta_{t+1}^\circ=\eta_t^\circ P_{\mu_t^\circ},
\qquad t\geq1,
\end{equation}
where $(\eta P_\pi)(D)=\int \limits_{\cS}P_\pi(s,D)\,d\eta(s)$. These distributions continue to evolve after the policy switch and need not be stationary.

Using these distributions, define the batch-$k$ population matrices by
\begin{align}
W_{\mathrm{cum},k}(\pi)
&=
\frac{1}{T_k}\sum \limits_{t=1}^{T_k}
\int \limits_{\cS}\sum \limits_{a\in\cA}
\pi(a;s)\Phi(s)
\{\Phi(s)-\gamma(P_a\Phi)(s)\}^\top
\,d\eta_t^\circ(s),
\label{eq:population-W}\\
b_{\mathrm{cum},k}(\pi)
&=
\frac{1}{T_k}\sum \limits_{t=1}^{T_k}
\int \limits_{\cS}\sum \limits_{a\in\cA}
\pi(a;s)r_a(s)\Phi(s)
\,d\eta_t^\circ(s).
\label{eq:population-b}
\end{align}

Under Assumption~\ref{ass:specification} below, these matrices nevertheless identify the same coefficient $\theta_0^\pi$ for every $k$, as shown in Lemma~\ref{lem:population}.

\begin{assumption}[Batch sizes]\label{ass:batch}
The sequence $\{B_k\}_{k\geq1}$ is deterministic and satisfies $0<B_{\min}\leq B_k\leq B_{\max}<\infty$ for all $k\geq1$.
\end{assumption}

\begin{assumption}[Treatment assignment and conditional independence]\label{ass:randomization}

The initial states $S_1^1,\ldots,S_1^n$ are independent and identically distributed with common distribution $\eta_1^\circ$. Let
\[
\mathcal G_{t-1}
=
\sigma\bigl(
S_1^1,\ldots,S_1^n,\,
A_s^i,U_s^i,S_{s+1}^i:
1\le i\le n,\ 1\le s<t
\bigr)
\]
denote the information collected through time $t-1$, including the current states $S_t^1,\ldots,S_t^n$. Let $\mu_{n,t}$ be the behavior policy selected at the preceding batch boundary, with $\mu_{n,t}=\pi^0$ in the first batch. Define
\[
\mathcal F_{t,0}
=
\mathcal G_{t-1}
\vee
\sigma(\mu_{n,1},\ldots,\mu_{n,t}),
\]
and, for $i=1,\ldots,n$,
\[
\mathcal F_{t,i}
=
\mathcal F_{t,0}
\vee
\sigma\bigl(
A_t^\ell,U_t^\ell,S_{t+1}^\ell:
1\le\ell\le i
\bigr).
\]
Thus, $\mathcal F_{t,0}$ contains all information available before treatment assignment at time $t$, including previously implemented policies, and
\[
\mathcal F_{t,n}\subseteq\mathcal F_{t+1,0}.
\]
Conditional on $\cF_{t,0}$, the pairs $(A_t^i,S_{t+1}^i)$, $i=1,\ldots,n$, are independent and satisfy
\[
\P(A_t^i=a,S_{t+1}^i\in D\mid\cF_{t,0})
=\mu_{n,t}(a;S_t^i)P_a(S_t^i,D).
\]
The transition kernels $P_a$ do not depend on $t$, $i$, or $n$. The utility is $U_t^i=u(S_{t+1}^i,A_t^i,S_t^i)$.

\end{assumption}

\begin{assumption}[State-space and transition regularity]\label{ass:state}
The state space $\cS\subset\R^p$ is compact; $\Phi$ and $u$ are continuous on their respective domains. For each $a\in\cA$ and every bounded continuous $f:\cS\to\R$, the function
\[
s\longmapsto(P_af)(s)=\int \limits_{\cS}f(s')P_a(s,ds')
\]
is continuous. The known initial policy $\pi^0(a;s)$ is continuous and strictly positive for each action $a$ and every state $s\in\cS$.
\end{assumption}

\begin{assumption}[Linear value functions]\label{ass:specification}
For every $\pi\in\Pi$, there exists $\theta_0^\pi\in\operatorname{int}(\Theta)$ such that $V^\pi(s)=\Phi(s)^\top\theta_0^\pi$ for every $s\in\cS$. The set $\Theta\subset\R^q$ is closed and convex.
\end{assumption}

Assumption~\ref{ass:specification} imposes exact linear specification of the value functions over the candidate policy class. It ensures that the importance-weighted Bellman estimating equation has conditional mean zero under any admissible behavior policy, a property proved later and used throughout the analysis.

\begin{assumption}[Identifiability]\label{ass:identifiability}

The matrix $\Omega$ is symmetric positive definite. For each fixed $k\geq1$, there exists $c_{1,k}>0$ such that
\[
\inf_{\beta\in\mathcal B}
\sigma_{\min}\left\{
W_{\mathrm{cum},k}(\pi_\beta)
\right\}
\geq c_{1,k},
\]
where $\sigma_{\min}(A)$ denotes the smallest singular value of $A$.

\end{assumption}

\begin{assumption}[Penalty]\label{ass:penalty}
The penalty $J:\Theta\to[0,\infty)$ is continuous and convex. It is continuously differentiable on an open neighborhood of $\theta_0$, with derivative bounded on compact subsets of that neighborhood. 
For each fixed $k\geq1$, assume
\[
\lambda_{n,k}=o(N_k^{-1/2})
\qquad\text{as }n\to\infty,
\]
and, for each fixed $n\geq1$,
\[
\lambda_{n,k}\to0
\qquad\text{as }k\to\infty.
\]

\end{assumption}

\begin{assumption}[Unique population optimum]\label{ass:unique}
The function $M$ defined in \eqref{eq:population-optimum} has a unique maximizer $\beta^*\in\cB$.
\end{assumption}

\begin{remark}\label{rem:positivity}
Assumption~\ref{ass:state}, compactness of $\cS$ and $\cB$, and the logistic form of $\pi_\beta$ imply that there is $c_0>0$ for which
\[
\inf_{n,t,a,s}\mu_{n,t}(a;s)\geq c_0,\qquad
\inf_{t,a,s}\mu_t^\circ(a;s)\geq c_0
\]
almost surely. 
Consequently, the importance weights are uniformly bounded, as are the rewards and features by compactness and continuity. We do not assume that the state process is stationary following a policy switch.

\end{remark}

\begin{lemma}[Population equation and continuity]\label{lem:population}
Under Assumptions~\ref{ass:state}--\ref{ass:identifiability}, for all $\pi\in\Pi$ and all $k\geq1$,
\begin{equation}\label{eq:common-target}
b_{\mathrm{cum},k}(\pi)=W_{\mathrm{cum},k}(\pi)\theta_0^\pi.
\end{equation}
Moreover, $\beta\mapsto\theta_0^{\pi_\beta}$ and $M(\beta)=\nu^\top\theta_0^{\pi_\beta}$ are bounded and Lipschitz continuous on $\cB$.
\end{lemma}

The proof is given in Appendix~\ref{prf:lem3.1}. Thus the batch sizes and behavior policies change the population matrices but not their solution: adaptivity affects how precisely $\theta_0^\pi$ and $\beta^*$ are estimated, not what they are.

\section{Asymptotics for fixed number of batches}\label{sec:fixed-batch}

We first study the regime in which many patients are enrolled and the policy is updated only a few times. Formally, we take $n\to\infty$ with the batch index $k$ fixed, so that the time horizon $T_k$ is fixed while the number of observed patient histories grows. Because each batch's policy is estimated from pooled data, these histories are dependent and their distribution changes across batches. 

For each patient $i$ and time $t$, let
\[
H_{i,t}=(S_1^i,A_1^i,S_2^i,\ldots,A_{t-1}^i,S_t^i),
\qquad
\mathbb Q_{n,t}=\frac1n\sum \limits_{i=1}^n\delta_{H_{i,t}}.
\]
The history $H_{i,t}$ takes values in the compact space $\cH_t=\cS^t\times\cA^{t-1}$.

Let $Q_1=\eta_1^\circ$, and define $Q_t$ recursively as the distribution of $H_t$ under the limiting behavior-policy sequence. Given $H_t=h$, draw $A_t$ according to $\mu_t^\circ(\cdot;S_t)$ and $S_{t+1}$ according to $P_{A_t}(S_t,\cdot)$, and append $(A_t,S_{t+1})$ to obtain $H_{t+1}$. The utility $U_t=u(S_{t+1},A_t,S_t)$ is determined by these variables. By construction, the marginal distribution of $S_t$ under $Q_t$ is $\eta_t^\circ$.

We write
\[
d_{\mathrm{BL}}(Q,Q')=\sup_{\|f\|_\infty\leq1,\,\operatorname{Lip}(f)\leq1}
\left|\int f\,dQ-\int f\,dQ'\right|
\]
for the bounded-Lipschitz distance on a compact history space. This metric generates weak convergence of probability measures on that space.

\subsection{Consistency}

Our first result establishes consistency of the cumulative V-learning estimator for a fixed number of batches as the number of patients increases. We first establish convergence of their empirical distributions and the corresponding estimating matrices, and then use the common population equation to obtain consistency of the value-function and policy estimators.

\begin{theorem}[Consistency]\label{thm:consistency}
Under Assumptions~\ref{ass:batch}--\ref{ass:unique}, for every fixed $k\geq1$, as $n\to\infty$, the following hold:
\begin{enumerate}[label=(\roman*)]
\item \textit{Convergence of empirical history distributions.} For each $t=1,\ldots,T_k+1$,
\[
d_{\mathrm{BL}}(\mathbb Q_{n,t},Q_t)\xrightarrow{\mathbb{P}}0.
\]
In particular, the empirical state distribution at time $t$ converges weakly in probability to $\eta_t^\circ$.

\item \textit{Uniform convergence of the estimating matrices.}
\begin{align}
\sup_{\pi\in\Pi}
\|\hat W_{\mathrm{cum},k}(\pi)-W_{\mathrm{cum},k}(\pi)\|
&\xrightarrow{\mathbb{P}}0,\nonumber\\
\sup_{\pi\in\Pi}
\|\hat b_{\mathrm{cum},k}(\pi)-b_{\mathrm{cum},k}(\pi)\|
&\xrightarrow{\mathbb{P}}0.
\label{eq:matrix-consistency}
\end{align}

\item \textit{Consistency of the value-function and policy estimators.}
\begin{align}
\sup_{\pi\in\Pi}
\|\hat\theta_{\pi,n}^k-\theta_0^\pi\|
&\xrightarrow{\mathbb{P}}0,\nonumber\\
\hat\beta_n^k&\xrightarrow{\mathbb{P}}\beta^*.
\label{eq:uniform-consistency}
\end{align}
\end{enumerate}
\end{theorem}

The proof, given in Appendix~\ref{prf:thm4.1}, proceeds by induction over batches. The result shows that, although the behavior policies and state distributions evolve across batches, the cumulative estimator consistently identifies the same population value-function parameters and optimal policy. 

\subsection{Asymptotic normality and oracle equivalence}
For $\beta\in\cB$ abbreviate $\theta_\beta=\theta_0^{\pi_\beta}$ and define
\begin{equation}\label{eq:score}
\xi_t^i(\beta)=\frac{\pi_\beta(A_t^i;S_t^i)}{\mu_{n,t}(A_t^i;S_t^i)}
e_t^i(\theta_\beta)\Phi(S_t^i),\qquad
\mathbb G_{n,k}(\beta)=\frac1{\sqrt{N_k}}\sum \limits_{t=1}^{T_k}\sum \limits_{i=1}^n\xi_t^i(\beta).
\end{equation}

With the filtration $\{\cF_{t,i}\}$ defined in Assumption~\ref{ass:randomization}, the conditional treatment assignment mechanism and the Bellman equation give
\begin{equation}\label{eq:martingale-centering}
\E[\xi_t^i(\beta)\mid\cF_{t,i-1}]
=\Phi(S_t^i)\sum \limits_{a\in\cA}\pi_\beta(a;S_t^i)
\{r_a(S_t^i)+\gamma(P_a\Phi)(S_t^i)^\top\theta_\beta-\Phi(S_t^i)^\top\theta_\beta\}=0.
\end{equation}

Thus, the summands in \eqref{eq:score} are martingale differences. 

For $\beta,\beta'\in\cB$ define
\begin{align*}
g_t(\beta,\beta';s)
&=\sum \limits_{a\in\cA}\frac{\pi_\beta(a;s)\pi_{\beta'}(a;s)}{\mu_t^\circ(a;s)}\Phi(s)\Phi(s)^\top\\
&\quad\times\int \limits_{\cS}\left\{u(s',a,s)+\gamma\Phi(s')^\top\theta_\beta-\Phi(s)^\top\theta_\beta\right\}\\
&\qquad\qquad\times\left\{u(s',a,s)+\gamma\Phi(s')^\top\theta_{\beta'}-\Phi(s)^\top\theta_{\beta'}\right\}
P_a(s,ds'),
\end{align*}
and set
\begin{equation}\label{eq:score-covariance}
\Gamma_k(\beta,\beta')=\frac1{T_k}\sum \limits_{t=1}^{T_k}\int \limits_{\cS}g_t(\beta,\beta';s)\,d\eta_t^\circ(s),\qquad
\Sigma_k(\beta,\beta')=W_{\mathrm{cum},k}(\pi_\beta)^{-1}\Gamma_k(\beta,\beta')W_{\mathrm{cum},k}(\pi_{\beta'})^{-\top}.
\end{equation}

\begin{lemma}[Stochastic equicontinuity of the Bellman scores]\label{lem:equicontinuity}
Under Assumptions~\ref{ass:batch}--\ref{ass:identifiability}, for each fixed $k\ge1$ and $\epsilon>0$,
\[
\lim_{\delta\downarrow0}\sup_{n\ge1}\P\!\left\{
\sup_{\substack{\beta,\beta'\in\cB\\\|\beta-\beta'\|\le\delta}}
\|\mathbb G_{n,k}(\beta)-\mathbb G_{n,k}(\beta')\|>\epsilon\right\}=0.
\]
\end{lemma}

The proof is given in Appendix~\ref{prf:lem4.2}. We next strengthen consistency to a functional central limit theorem, uniform over the policy class, which is needed for inference at the data-dependent $\hat\beta_n^k$.

\begin{theorem}[Functional asymptotic normality]\label{thm:functional-clt}
Under Assumptions~\ref{ass:batch}--\ref{ass:unique}, for each fixed $k\geq1$,
\begin{equation}\label{eq:functional-clt}
\left\{\sqrt{N_k}(\hat\theta_{\pi_\beta,n}^k-\theta_0^{\pi_\beta}):\beta\in\cB\right\}
\rightsquigarrow\{\mathbb Z_k(\beta):\beta\in\cB\}
\quad\text{in }\ell^\infty(\cB)^q,
\end{equation}
where $\mathbb Z_k$ is a tight, mean-zero Gaussian process with covariance
\[
\Cov\{\mathbb Z_k(\beta),\mathbb Z_k(\beta')\}
=\Sigma_k(\beta,\beta').
\]
Consequently, the process $\sqrt{N_k}\{\hat V_{n,R,k}(\pi_\beta)-M(\beta)\}$ converges weakly in $\ell^\infty(\cB)$ to $\nu^\top\mathbb Z_k(\beta)$.
\end{theorem}

The proof is given in Appendix~\ref{prf:thm4.3}. Although the behavior policies are estimated from earlier batches, $\Sigma_k$ depends on them only through their limits $\mu_t^\circ$, and their sampling variability adds no further term to the covariance. This is because the Bellman scores have conditional mean zero under every behavior policy, so errors in the estimated policies do not propagate to first order. In the terminology of \citet{ZhangJansonMurphy2022}, the estimating equation is policy invariant, and no adaptive correction to the variance is needed.


\begin{corollary}[First-order oracle equivalence]\label{cor:oracle}
Fix $k\ge1$ and let $\hat\theta_{\pi,n}^{k,\circ}$ denote the cumulative V-learning estimator computed from $n$ independent patient trajectories generated under $\mu_t^\circ$, using this policy in the importance-weight denominators and the same $\lambda_{n,k},J,\Phi,\Omega$. Under Assumptions~\ref{ass:batch}--\ref{ass:unique},
\[
\left\{\sqrt{N_k}(\hat\theta_{\pi_\beta,n}^{k,\circ}-\theta_0^{\pi_\beta}):\beta\in\cB\right\}
\ \rightsquigarrow\ \mathbb Z_k
\]
in $\ell^\infty(\cB)^q$, with the same Gaussian process $\mathbb Z_k$ and the same covariance $\Sigma_k$ as in Theorem~\ref{thm:functional-clt}. The corresponding value processes also have the same limit.
\end{corollary}

The proof is given in Appendix~\ref{prf:cor4.4}. 

\subsection{Inference for learned policy}
\begin{theorem}[Inference for the learned policy]\label{thm:policy-value}

Under Assumptions~\ref{ass:batch}--\ref{ass:unique}, for each fixed $k\geq1$, let
\[
\sigma_{V,k}^2=\nu^\top\Sigma_k(\beta^*,\beta^*)\nu.
\]
Then
\begin{equation}\label{eq:optimal-value-clt}
\sqrt{N_k}\{\hat V_{n,R,k}(\pi_{\hat\beta_n^k})-M(\hat\beta_n^k)\}
\xrightarrow{d}N(0,\sigma_{V,k}^2),
\end{equation}
and the same limit holds with $M(\hat\beta_n^k)$ replaced by $M(\beta^*)$.
\end{theorem}

The proof is given in Appendix~\ref{prf:thm4.5}. The theorem allows two centerings. The first, $M(\hat\beta^{k}_{n})$, is random: it is the value of the policy actually deployed after batch $k$, which is the quantity of practical interest and the standard target in inference for data-driven treatment regimes \citep{ChakrabortyLaberZhao2014}. The second, $M(\beta^{*})$, is the optimal value in the policy class. Since both centerings yield the same limit, the resulting interval is asymptotically valid for either target.\\

For $\beta\in\cB$, define the estimated patient-level score
\[
\hat\psi_i^k(\beta)=\frac1{\sqrt{T_k}}\sum \limits_{t=1}^{T_k}
\frac{\pi_\beta(A_t^i;S_t^i)}{\mu_{n,t}(A_t^i;S_t^i)}
e_t^i(\hat\theta_{\pi_\beta,n}^k)\Phi(S_t^i),
\]
and put
\begin{equation}\label{eq:sandwich}
\hat\Gamma_k(\beta,\beta')=\frac1n\sum \limits_{i=1}^n
\hat\psi_i^k(\beta)\hat\psi_i^k(\beta')^\top,
\qquad
\hat\Sigma_k(\beta,\beta')=
\hat W_{\mathrm{cum},k}(\pi_\beta)^{-1}\hat\Gamma_k(\beta,\beta')
\hat W_{\mathrm{cum},k}(\pi_{\beta'})^{-\top}.
\end{equation}
The inverse is understood on the event where the empirical matrices are nonsingular, which has probability tending to one.

\begin{proposition}[Consistency of the estimator]\label{prop:variance}
Under Assumptions~\ref{ass:batch}--\ref{ass:unique}, for each fixed $k\geq1$,
\[
\sup_{\beta,\beta'\in\cB}
\|\hat\Sigma_k(\beta,\beta')-\Sigma_k(\beta,\beta')\|\xrightarrow{\mathbb{P}}0.
\]
Consequently, $\hat\sigma_{V,k}^2=\nu^\top\hat\Sigma_k(\hat\beta_n^k,\hat\beta_n^k)\nu\to\sigma_{V,k}^2$ in probability. If $\sigma_{V,k}^2>0$, the statistic in \eqref{eq:optimal-value-clt} divided by $\hat\sigma_{V,k}$ converges to a standard normal distribution.
\end{proposition}

The proof is given in Appendix~\ref{prf:prop4.6}.

\section{Asymptotics for fixed number of patients}\label{sec:fixed-n}

We now consider the complementary regime, in which a small cohort is followed over many batches, as in a study with few participants but long follow-up. Formally, we fix $n$ and let $k\to\infty$. Under Assumption~\ref{ass:batch}, $T_k\to\infty$, so the cumulative number of observations $N_k=nT_k$ also diverges. Information now accrues only through repeated decisions within patients, so consistency must come from the length of follow-up rather than the size of the cohort.

For $\beta\in\cB$, define
\begin{equation}\label{eq:predictable-W}
\overline W_{n,k}(\pi_\beta)
=
\frac1{N_k}\sum \limits_{t=1}^{T_k}\sum \limits_{i=1}^n
\sum \limits_{a\in\cA}\pi_\beta(a;S_t^i)\Phi(S_t^i)
\{\Phi(S_t^i)-\gamma(P_a\Phi)(S_t^i)\}^\top.
\end{equation}

\begin{assumption}[Persistent identifiability]\label{ass:persistent}
For each fixed $n\geq1$,
\[
\liminf_{k\to\infty}
\inf_{\beta\in\cB}
\sigma_{\min}\{\overline W_{n,k}(\pi_\beta)\}>0
\text{ almost surely},
\]
where $\sigma_{\min}(A)$ denotes the smallest singular value of $A$.
\end{assumption}

Assumption~\ref{ass:persistent} requires the realized trajectory to keep exploring the state space, analogous to persistent-excitation conditions in adaptive regression \citep{LaiWei1982}. The following lemma provides the uniform strong law needed for both the empirical estimating matrices and the Bellman scores.

\begin{lemma}[Uniform strong law for martingale differences]\label{lem:longrun-slln}
Let $\{X_t(\beta):\beta\in\cB\}$, $t\geq1$, be $\R^d$-valued random functions, where $d$ is fixed and $\cB$ is compact. Suppose that $X_t(\beta)$ is $\cF_{t+1,0}$-measurable and
\[
\E[X_t(\beta)\mid\cF_{t,0}]=0
\]
for every $t$ and $\beta$. Suppose also that there exist deterministic constants $C,L<\infty$ such that, almost surely, for all $t$ and $\beta,\beta'\in\cB$,
\[
\|X_t(\beta)\|\leq C,\qquad
\|X_t(\beta)-X_t(\beta')\|
\leq L\|\beta-\beta'\|.
\]
Then
\[
\sup_{\beta\in\cB}
\left\|\frac1T\sum \limits_{t=1}^TX_t(\beta)\right\|
\xrightarrow{\mathrm{a.s.}}0
\qquad\text{as }T\to\infty.
\]
\end{lemma}

The proof is given in Appendix~\ref{prf:lem5.1}. 

Many mHealth studies follow a small cohort over a long period. 
The next theorem shows that consistency is possible with a fixed cohort, regardless of its size, provided the persistent identifiability condition holds.

\begin{theorem}[Consistency with a fixed number of patients]\label{thm:longrun-consistency}
Suppose Assumptions~\ref{ass:batch}--\ref{ass:persistent} hold and the online procedure is well-defined at each batch. Then, for each fixed $n\geq1$, the following hold as $k\to\infty$:

\begin{enumerate}[label=(\roman*)]
\item The empirical estimating matrices satisfy
\begin{equation}\label{eq:longrun-matrix}
\sup_{\beta\in\cB}
\|\hat W_{\mathrm{cum},k}(\pi_\beta)
-\overline W_{n,k}(\pi_\beta)\|
\xrightarrow{\mathrm{a.s.}}0.
\end{equation}
\item The cumulative value-function estimators are uniformly consistent:
\begin{equation}\label{eq:longrun-theta}
\sup_{\beta\in\cB}
\|\hat\theta_{\pi_\beta,n}^k-\theta_0^{\pi_\beta}\|
\xrightarrow{\mathrm{a.s.}}0.
\end{equation}
\item The estimated policy parameter satisfies
\begin{equation}\label{eq:longrun-beta}
\hat\beta_n^k\xrightarrow{\mathrm{a.s.}}\beta^*.
\end{equation}
\end{enumerate}
\end{theorem}

The proof is given in Appendix~\ref{prf:thm5.2}.

\section{Simulation studies}\label{sec:simulations}

We simulate online V-learning in two experiments, each with 100 replications: the first fixes the number of batches and increases the number of patients, to assess value estimation and inference; the second experiment fixes the number of patients and follows the estimated policies as batches accumulate.

\subsection{Simulation Setup}

We use a generative model similar to that of \citet{Luckett2020}. Let $S_t=(S_{t,1},S_{t,2})^\top$ denote the state and $A_t\in\{0,1\}$ the treatment at time $t$. Suppressing the patient index, we generate
\begin{align}
X_{t+1,1}&=0.75(2A_t-1)S_{t,1}+0.25S_{t,1}S_{t,2}+\epsilon_{t,1},\nonumber\\
X_{t+1,2}&=0.75(1-2A_t)S_{t,2}+0.25S_{t,1}S_{t,2}+\epsilon_{t,2},
\label{eq:sim-transition}
\end{align}
where the $\epsilon_{t,j}$ are independent $N(0,1/4)$ random variables across patients, times, and coordinates, and set \[S_{t+1,j}=\max\{-6,\min(6,X_{t+1,j})\}\]
for $j=1,2$ to prevent numerical instability. The utility \[U_t=2S_{t+1,1}+S_{t+1,2}-0.25(2A_t-1)\] 
rewards larger state values and imposes a cost for treatment. Initial states are independent standard normal, and observations are recorded after a burn-in of 50 time points under uniform treatment assignment.

We take $\gamma=0.9$ and the randomized policies \[\pi_\beta(1;s)=\exp(s^\top\beta)/\{1+\exp(s^\top\beta)\}, \qquad \beta\in[-5,5]^2.\] Following \citet{Luckett2020}, we model the state-value function with Gaussian radial basis functions $\phi(z;c,\tau)=\exp\{-(z-c)^2/(2\tau^2)\}$, with $\tau=0.25$ and $c\in\{0,0.25,0.5,0.75,1\}$, applied separately to each state coordinate. Let $\{s_j^R\}_{j=1}^{m_R}$ be a reference sample of $m_R=5{,}000$ states. Each coordinate is first mapped to $[0,1]$ using the first and ninety-ninth percentiles of this sample, with values outside this interval truncated. Together with an intercept, this gives a basis $\widetilde\Phi$ of 11 functions.

Because the five Gaussian functions for each coordinate nearly sum to a constant on $[0,1]$, $\widetilde\Phi$ is close to collinear with the intercept: its Gram matrix $G_R=m_R^{-1}\sum \limits_{j=1}^{m_R}\widetilde\Phi(s_j^R)\widetilde\Phi(s_j^R)^\top$ over the reference sample has condition number of order $10^5$. We therefore use the whitened basis $\Phi(s)=G_R^{-1/2}\widetilde\Phi(s)$ \citep{KessyLewinStrimmer2018}, held fixed across batches and replications. Whitening preserves the linear span of the basis and improves numerical conditioning. The ridge penalty, however, is applied in the whitened coordinates and is not invariant to this reparametrization \citep{GolubVanLoan2013}. We take $\Omega=I_{11}$, $J(\theta)=\|\theta\|^2$ and $\lambda_{n,k}=0.01/N_k$.

The first batch contains 12 decision times per patient under $\pi^0(1;s)=1/2$, and each later batch contains six, so that $T_k=6k+6$ for $k\ge1$. After each batch, the policy is re-estimated from all available data and deployed during the next batch. The same patients are followed throughout, and their states are not reset when the policy changes.

As in \citet{Luckett2020}, we add an $L_2$ penalty on the policy coefficients, so that at batch $k$ the estimated policy maximizes 
\[\hat M_{n,k}(\beta)=\hat V_{n,R,k}(\pi_\beta)-\rho\|\beta\|^2\]
with $\rho=0.5$. The penalty keeps the optimum away from the boundary of the parameter space, where policies are nearly deterministic and importance weights become extreme. Because it is used only in the simulations, the target is the maximizer of $M_\rho(\beta)=V_R(\pi_\beta)-\rho\|\beta\|^2$ rather than $\beta^*$ in \eqref{eq:population-optimum}.

We compute $\hat\beta_n^k$ by maximizing $\hat M_{n,k}$ over $[-5,5]^2$ with the L-BFGS-B algorithm, as implemented in the \texttt{optim} function in R. The starting values are the zero vector, eight points drawn uniformly from $[-5,5]^2$ and, after the first batch, the preceding estimate $\hat\beta_n^{k-1}$. Candidate policies whose empirical estimating matrix is singular or has condition number greater than 200 are excluded; if no admissible solution is found, the preceding policy is retained.

The reference distribution $R$ is the empirical distribution of the $m_R=5{,}000$ reference states, generated by the initialization procedure above. This sample is held fixed across replications and is also used to compute $\nu$. We approximate $V_R(\pi_\beta)$ by averaging discounted utilities over 5,000 trajectories of length 120 started at the reference states, and write $\tilde V_R(\pi_\beta)$ and $\tilde M_\rho(\beta)=\tilde V_R(\pi_\beta)-\rho\|\beta\|^2$ for the resulting approximations. The maximizer of $M_\rho$ is located by grid search with successive spacings $0.5$, $0.1$ and $0.025$. The 25 highest-ranked candidates are then re-evaluated on independent samples, the value of the selected policy is reported from a separate Monte Carlo sample, and common random numbers are used when comparing policies. 

The resulting benchmark is $\tilde\beta^*_\rho=(1.550,-0.925)^\top$, with $\tilde V_R(\pi_{\tilde\beta^*_\rho})=5.941$ (Monte Carlo standard error $0.061$) and $\tilde M_\rho(\tilde\beta^*_\rho)=4.312$.
\paragraph{Fixed number of batches.}
We take $n\in\{25,100,250\}$ and evaluate the estimators after batches $k\in\{1,3,5\}$. For comparison, we consider an oracle experiment in which the first batch is generated under $\pi^0$ and all later batches under the benchmark policy $\pi_{\tilde\beta_\rho^*}$. The policy is re-estimated after each batch in both experiments, but under the oracle scheme treatment assignment continues to follow the benchmark policy.

Let $\hat\pi_n^k=\pi_{\hat\beta_n^k}$. We examine the estimation error $\hat V_{n,R,k}(\hat\pi_n^k)-\tilde V_R(\hat\pi_n^k)$ and nominal 95\% confidence intervals $\hat V_{n,R,k}(\hat\pi_n^k)\pm 1.96\,\hat\sigma_{V,k}/\sqrt{N_k}$, where $\hat\sigma_{V,k}^2$ is the variance estimator. Table~\ref{tab:simulation-inference} reports the empirical bias and standard deviation (SD) of the estimation error, the average estimated standard error (SE), and the coverage probability (CP) over 100 replications. Coverage is assessed relative to $\tilde V_R(\hat\pi_n^k)$, whose Monte Carlo error is not reflected in the estimated standard errors.

\begin{table}[!htbp]
\centering
\footnotesize
\setlength{\tabcolsep}{4.5pt}
\begin{tabular}{rr|rrrr|rrrr}
\hline
& & \multicolumn{4}{c|}{Adaptive} & \multicolumn{4}{c}{Oracle}\\
$n$ & $k$ & Bias & SD & SE & CP (\%) & Bias & SD & SE & CP (\%)\\
\hline
25  & 1 &  0.481 & 2.253 & 1.646 & 92 & 0.937 & 2.936 & 2.070 & 96\\
    & 3 &  0.059 & 0.785 & 0.758 & 94 & 0.225 & 0.830 & 0.787 & 95\\
    & 5 &  0.038 & 0.575 & 0.589 & 96 & 0.139 & 0.663 & 0.615 & 91\\
\hline
100 & 1 & -0.020 & 0.640 & 0.608 & 94 & 0.050 & 0.689 & 0.619 & 91\\
    & 3 &  0.011 & 0.400 & 0.384 & 92 & 0.052 & 0.386 & 0.396 & 98\\
    & 5 & -0.003 & 0.298 & 0.304 & 96 & 0.049 & 0.306 & 0.310 & 97\\
\hline
250 & 1 &  0.030 & 0.388 & 0.382 & 95 & 0.031 & 0.409 & 0.386 & 95\\
    & 3 &  0.062 & 0.284 & 0.245 & 93 & 0.049 & 0.255 & 0.248 & 97\\
    & 5 &  0.058 & 0.201 & 0.194 & 94 & 0.046 & 0.203 & 0.196 & 98\\
\hline
\end{tabular}
\caption{Estimation of the value of the estimated policy under adaptive and oracle treatment assignment: empirical bias and standard deviation (SD) of the estimation error, average estimated standard error (SE), and coverage probability (CP) of nominal 95\% confidence intervals, based on 100 replications per setting.}
\label{tab:simulation-inference}
\end{table}

Under both schemes, the standard deviations decrease as $n$ and $k$ increase, and the estimated standard errors are close to them except in the smallest settings, where the largest biases also occur. Coverage ranges from 92\% to 96\% under adaptive sampling and from 91\% to 98\% under oracle sampling, consistent with nominal coverage given a Monte Carlo standard error of about 2.2 percentage points.

\begin{figure}[!htbp]
\centering
\includegraphics[width=\linewidth]{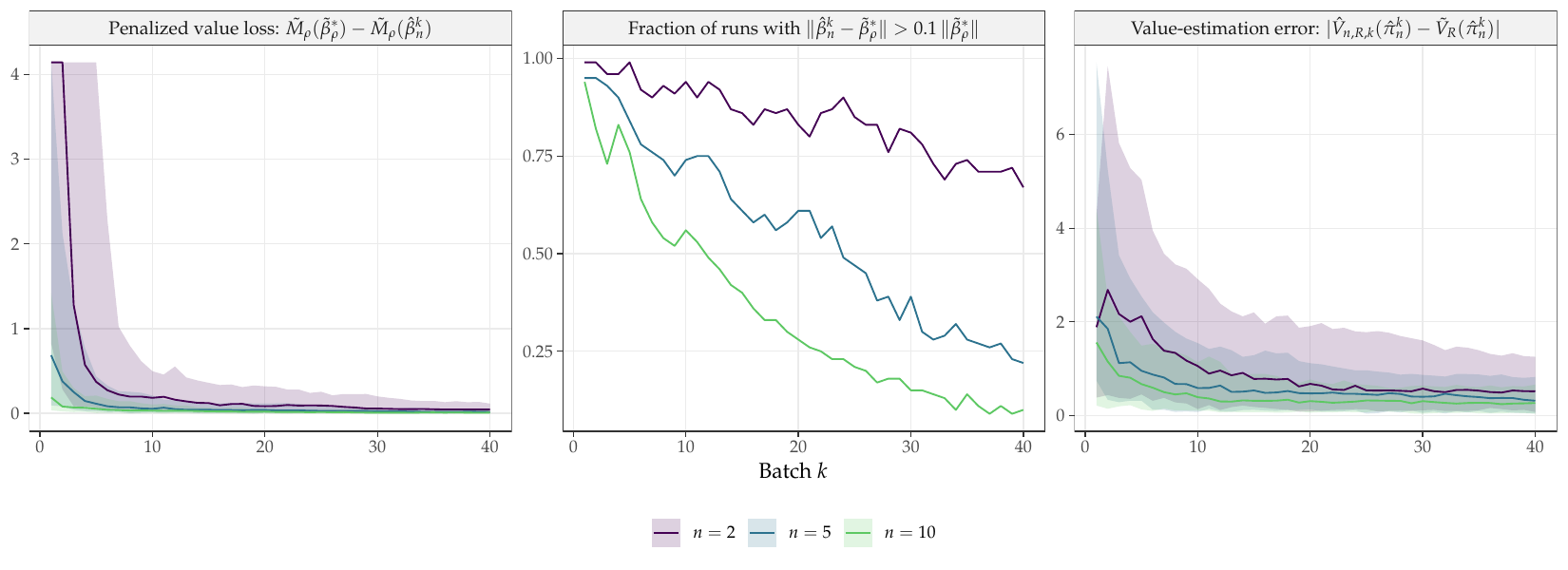}
\caption{Online V-learning with $n=2,5,10$ patients over 40 batches. (a)~Loss in penalized value $L_{n,k}$ and (c)~error in estimated value $D_{n,k}$: median (line) and 10th to 90th percentiles (band) across replications. (b)~Proportion of replications with $I_{n,k}=1$.}
\label{fig:simulation-longrun}
\end{figure}

At batch 5, the standard deviations under the two schemes become increasingly similar as $n$ grows, in line with first-order oracle equivalence. For $n=250$, the adaptive biases remain between $0.030$ and $0.062$; they are small relative to the standard deviations and may reflect basis approximation error as well as finite-sample and Monte Carlo effects.
\paragraph{Fixed number of patients.}
We next take $n\in\{2,5,10\}$ and follow each patient through 40 batches, or 246 decision times. After each batch we record the loss in penalized value $(L_{n, k})$, whether the policy parameter is off by more than 10\% ($I_{n, k}$), and the error in the estimated value ($D_{n, k}$) defined as:
\begin{align}
L_{n,k}&=\tilde M_\rho(\tilde\beta_\rho^*)-\tilde M_\rho(\hat\beta_n^k),\nonumber\\
I_{n,k}&=\mathbb{I}\left\{\|\hat\beta_n^k-\tilde\beta_\rho^*\|>0.1\|\tilde\beta_\rho^*\|\right\},\nonumber\\
D_{n,k}&=\left|\hat V_{n,R,k}(\hat\pi_n^k)-\tilde V_R(\hat\pi_n^k)\right|.
\label{eq:simulation-errors}
\end{align}
Figure~\ref{fig:simulation-longrun} summarizes these quantities across replications.

Figure~\ref{fig:simulation-longrun} shows that following a cohort for longer improves the estimated policy. The median loss in penalized value, shown in panel~(a), decreases with $k$ for all three cohort sizes, reaching $0.045$, $0.015$ and $0.008$ at batch 40 for $n=2$, $5$ and $10$. The proportion of replications with parameter error above 10\%, shown in panel~(b), also decreases, to 67\%, 22\% and 10\%, although slowly for $n=2$; there, the loss is small even when the parameter error remains large, so a near-optimal value does not require an accurate estimate of $\beta$. The median error in estimated value, shown in panel~(c), decreases to $0.516$, $0.315$ and $0.266$. Because the benchmark is computed by grid search and Monte Carlo simulation, losses can be negative, and small losses should be read in light of its precision.

Two further results are not shown in the figure. At batch 1, the estimating matrix is too poorly conditioned for a valid value estimate in 80\%, 17\% and 4\% of replications for $n=2$, $5$ and $10$; these replications are excluded from the summaries in panel~(c) only, and for $n=2$ the fallback rule accounts for part of the sharp initial decline in panel~(a). By batch 40, all replications yield valid estimates, with median condition numbers between $10.89$ and $12.15$.

\section{Discussion}\label{sec:discussion}

We have developed asymptotic theory for cumulative online V-learning with repeated policy updates from longitudinal data on a fixed cohort. For a fixed number of batches, the value-function estimators are uniformly consistent and satisfy a functional central limit theorem, with the same first-order limiting distribution as an oracle procedure. Inference for the value of the learned policy can be based on a patient-level variance estimator. With the number of patients fixed, consistency continues to hold as the number of batches increases under persistent identifiability.

A central limitation is the exact linear specification of the candidate-policy value functions, which ensures that the population target remains invariant under adaptive behavior policies. Understanding the effects of approximation error on this property and on subsequent inference is an important direction for future work. Several further directions remain. In the fixed-cohort regime we establish consistency but not a limiting distribution; inference there would require conditions under which the state process and the behavior policies stabilize over time, together with rates of convergence. A joint asymptotic theory, in which the numbers of patients and batches grow together, would connect the two regimes and clarify how the length of follow-up and the size of the cohort trade off. Extending the results to data-dependent batch lengths is another direction, particularly since adaptive update schedules are common in practice. Finally, our results assume a unique optimal policy with assignment probabilities bounded away from zero. When the optimal policy is nearly deterministic, importance weights can become large in finite samples, and when the optimum is not unique, the usual Gaussian limit for the estimated optimal value need not apply.
 Both situations arise naturally in practice and merit further study.

\appendix

\section{Proof for Section~\ref{sec:online}}\label{app:online}

\phantomsection\label{prf:lem3.1}
\begin{proof}[Proof of Lemma~\ref{lem:population}]
Fix $\pi\in\Pi$. Since $\Phi(s)^\top\theta_0^\pi$ is the value of $\pi$, the Bellman equation gives, for each $s\in\cS$,
\[
\Phi(s)^\top\theta_0^\pi
=\sum \limits_{a\in\cA}\pi(a;s)\{r_a(s)+\gamma(P_a\Phi)(s)^\top\theta_0^\pi\}.
\]
Multiplying by $\Phi(s)$ and rearranging yields
\[
\sum \limits_{a\in\cA}\pi(a;s)r_a(s)\Phi(s)
=\left[\sum \limits_{a\in\cA}\pi(a;s)\Phi(s)
\{\Phi(s)-\gamma(P_a\Phi)(s)\}^\top\right]\theta_0^\pi.
\]

Integrating with respect to $\eta_t^\circ$ and averaging over $t=1,\ldots,T_k$ yields
\begin{align*}
b_{\mathrm{cum},k}(\pi)
&=\frac{1}{T_k}\sum \limits_{t=1}^{T_k}
\int \limits_{\cS}\sum \limits_{a\in\cA}
\pi(a;s)r_a(s)\Phi(s)\,d\eta_t^\circ(s)\\
&=\left[
\frac{1}{T_k}\sum \limits_{t=1}^{T_k}
\int \limits_{\cS}\sum \limits_{a\in\cA}
\pi(a;s)\Phi(s)
\{\Phi(s)-\gamma(P_a\Phi)(s)\}^\top
\,d\eta_t^\circ(s)
\right]\theta_0^\pi\\
&=W_{\mathrm{cum},k}(\pi)\theta_0^\pi,
\end{align*}
which proves \eqref{eq:common-target}.

For $\beta\in\cB$, put $W_\beta=W_{\mathrm{cum},1}(\pi_\beta)$ and $b_\beta=b_{\mathrm{cum},1}(\pi_\beta)$. Assumption~\ref{ass:identifiability} implies
\[
\sup_{\beta\in\cB}\|W_\beta^{-1}\|\leq c_{1,1}^{-1}.
\]
By compactness, the boundedness of $u$ and $\Phi$, and \eqref{eq:population-b}, $C_b:=\sup_\beta\|b_\beta\|<\infty$. Therefore, by \eqref{eq:common-target},
\[
\sup_{\beta\in\cB}\|\theta_0^{\pi_\beta}\|
=\sup_\beta\|W_\beta^{-1}b_\beta\|\leq c_{1,1}^{-1}C_b.
\]
Because $\cS$ and $\cB$ are compact, the derivatives of the logistic probabilities with respect to $\beta$ are uniformly bounded. Thus, for some $L_\pi<\infty$,
\[
\sup_{s,a}|\pi_\beta(a;s)-\pi_{\beta'}(a;s)|\leq L_\pi\|\beta-\beta'\|.
\]
The definitions of $W_\beta,b_\beta$ consequently give finite constants $L_W,L_b$ for which
\[
\|W_\beta-W_{\beta'}\|\leq L_W\|\beta-\beta'\|,
\qquad\|b_\beta-b_{\beta'}\|\leq L_b\|\beta-\beta'\|.
\]
Using $W_\beta^{-1}-W_{\beta'}^{-1}=W_\beta^{-1}(W_{\beta'}-W_\beta)W_{\beta'}^{-1}$ gives
\begin{align*}
\|\theta_0^{\pi_\beta}-\theta_0^{\pi_{\beta'}}\|
&\leq\|W_\beta^{-1}(b_\beta-b_{\beta'})\|
+\|(W_\beta^{-1}-W_{\beta'}^{-1})b_{\beta'}\|\\
&\leq(c_{1,1}^{-1}L_b+c_{1,1}^{-2}C_bL_W)\|\beta-\beta'\|.
\end{align*}
Since $\nu$ is fixed, the same conclusion holds for $M(\beta)=\nu^\top\theta_0^{\pi_\beta}$.

\end{proof}

\section{Proofs for Section~\ref{sec:fixed-batch}}\label{app:fixed-batch}

\phantomsection\label{prf:thm4.1}
\begin{proof}[Proof of Theorem~\ref{thm:consistency}]
We establish (i)--(iii) jointly by induction on the batch index $j$. 

Since $H_{i,1}=S_1^i$, $i=1,\ldots,n$, are i.i.d.\ with law $Q_1$, the law of large numbers and a finite-net approximation of the bounded-Lipschitz unit ball, as used in the induction step below, give
\[
d_{\mathrm{BL}}(\mathbb Q_{n,1},Q_1)\xrightarrow{\mathrm{a.s.}}0.
\]

Suppose that, at the beginning of batch $j$,
\[
d_{\mathrm{BL}}(\mathbb Q_{n,T_{j-1}+1},
Q_{T_{j-1}+1})\xrightarrow{\mathbb{P}}0,
\]
and, if $j>1$, that $\hat\beta_n^{j-1}\xrightarrow{\mathbb{P}}\beta^*$. For $j=1$, the behavior policy is $\pi^0=\mu_t^\circ$. For $j>1$, the uniform Lipschitz continuity of the logistic policy class gives
\begin{equation}\label{eq:behavior-convergence}
\sup_{T_{j-1}<t\leq T_j}\sup_{a,s}
|\mu_{n,t}(a;s)-\mu_t^\circ(a;s)|
\xrightarrow{\mathbb{P}}0.
\end{equation}
We prove each assertion at batch $j$.

\begin{enumerate}[label=(\roman*)]
\item Fix $T_{j-1}<t\leq T_j$ and suppose
$d_{\mathrm{BL}}(\mathbb Q_{n,t},Q_t)\xrightarrow{\mathbb{P}}0$.
For $f\in C(\cH_{t+1})$, define
\[
(K_\mu f)(h)
=
\sum \limits_{a\in\cA}\mu(a;s_t)
\int \limits_{\cS}f(h,a,s')P_a(s_t,ds'),
\]
where $h=(s_1,a_1,\ldots,s_t)\in\cH_t$.

We show that $K_{\mu_t^\circ}f\in C(\cH_t)$. Let $h_m\to h$ in $\cH_t$, and denote their respective current states by $s_m$ and $s$, so that $s_m\to s$. For each $a\in\cA$, write $g_{m,a}(s')=f(h_m,a,s')$ and $g_a(s')=f(h,a,s')$. By continuity of $f$ and $u$ on the corresponding compact spaces,
\[
\|g_{m,a}-g_a\|_\infty\to0.
\]
Consequently,
\begin{align*}
&\left|\int g_{m,a}(s')P_a(s_m,ds')
-\int g_a(s')P_a(s,ds')\right|\\
&\qquad\leq
\|g_{m,a}-g_a\|_\infty
+\left|(P_ag_a)(s_m)-(P_ag_a)(s)\right|
\longrightarrow0,
\end{align*}
where the second term vanishes by Assumption~\ref{ass:state}, since $g_a\in C(\cS)$. Continuity of $\mu_t^\circ(a;\cdot)$ and finiteness of $\cA$ now imply $K_{\mu_t^\circ}f(h_m)\to K_{\mu_t^\circ}f(h)$, as required.

Moreover,
\[
\|K_{\mu_{n,t}}f-K_{\mu_t^\circ}f\|_\infty
\leq |\cA|\|f\|_\infty
\sup_{a,s}|\mu_{n,t}(a;s)-\mu_t^\circ(a;s)|
=o_{\mathbb{P}}(1).
\]

Conditional on $\cF_{t,0}$, the variables $f(H_{i,t+1})$ are independent, with conditional means $(K_{\mu_{n,t}}f)(H_{i,t})$. Consequently,
\[
\E[\mathbb Q_{n,t+1}f\mid\cF_{t,0}]
=\mathbb Q_{n,t}K_{\mu_{n,t}}f,
\qquad
\Var(\mathbb Q_{n,t+1}f\mid\cF_{t,0})
\leq\frac{\|f\|_\infty^2}{n}.
\]
Using $Q_{t+1}f=Q_tK_{\mu_t^\circ}f$, we obtain
\begin{align*}
\mathbb Q_{n,t+1}f-Q_{t+1}f
&=\{\mathbb Q_{n,t+1}f-
\mathbb Q_{n,t}K_{\mu_{n,t}}f\}\\
&\quad+\mathbb Q_{n,t}
(K_{\mu_{n,t}}-K_{\mu_t^\circ})f\\
&\quad+(\mathbb Q_{n,t}-Q_t)K_{\mu_t^\circ}f
=o_{\mathbb{P}}(1).
\end{align*}
The first term converges to zero by the conditional variance bound, the second by \eqref{eq:behavior-convergence}, and the third by the induction hypothesis and continuity of $K_{\mu_t^\circ}f$.

Let $\mathcal N_\epsilon$ be a finite $\epsilon$-net of the bounded-Lipschitz unit ball on $\cH_{t+1}$ in the supremum norm. Then
\[
d_{\mathrm{BL}}(\mathbb Q_{n,t+1},Q_{t+1})
\leq
\max_{f\in\mathcal N_\epsilon}
|(\mathbb Q_{n,t+1}-Q_{t+1})f|
+2\epsilon.
\]
The finite maximum is $o_{\mathbb{P}}(1)$, and $\epsilon$ is arbitrary. Iterating over the finitely many observation times in batch $j$ proves (i) through time $T_j+1$.

\item By the induction hypothesis and \eqref{eq:behavior-convergence},
\[
\max_{1\leq t\leq T_j}\sup_{a,s}
|\mu_{n,t}(a;s)-\mu_t^\circ(a;s)|
=o_{\mathbb{P}}(1).
\]

Let $c_0>0$ be the uniform lower bound on the behavior probabilities. For any $a\in\cA$ and $s\in\cS$,
\[
\left|
\frac{\pi_\beta(a;s)}{\mu_{n,t}(a;s)}
-\frac{\pi_\beta(a;s)}{\mu_t^\circ(a;s)}
\right|
\leq
c_0^{-2}|\mu_{n,t}(a;s)-\mu_t^\circ(a;s)|.
\]
Since $\Phi$ and $u$ are bounded, \eqref{eq:behavior-convergence} therefore implies that replacing $\mu_{n,t}$ by $\mu_t^\circ$ in the empirical matrices introduces an $o_{\mathbb P}(1)$ error, uniformly over $\beta\in\cB$.

To establish convergence of the resulting matrices, consider any one of their coordinates. It can be written as $\mathbb Q_{n,T_j+1}F_\beta$, where, for a complete history $
h=(s_1,a_1,s_2,\ldots,a_{T_j},s_{T_j+1})$,
\[
F_\beta(h)
=
\frac{1}{T_j}\sum \limits_{t=1}^{T_j}
\frac{\pi_\beta(a_t;s_t)}
{\mu_t^\circ(a_t;s_t)}G_t(h).
\]
Here $G_t(h)$ denotes the corresponding coordinate of either
$\Phi(s_t)\{\Phi(s_t)-\gamma\Phi(s_{t+1})\}^\top$
or $u(s_{t+1},a_t,s_t)\Phi(s_t)$, and is independent of $\beta$. By boundedness and continuity of the features, utility, and behavior probabilities, $F_\beta$ is bounded and continuous on $\cH_{T_j+1}$.

Moreover, there exists $C_G<\infty$ such that $\sup_{t,h}|G_t(h)|\leq C_G$. Since the logistic policy class satisfies
\[
\sup_{a,s}|\pi_\beta(a;s)-\pi_{\beta'}(a;s)|
\leq L_\pi\|\beta-\beta'\|,
\]
uniform positivity gives
\[
\sup_{h\in\cH_{T_j+1}}
|F_\beta(h)-F_{\beta'}(h)|
\leq \frac{C_G L_\pi}{c_0}
\|\beta-\beta'\|.
\]
Thus the required Lipschitz bound holds with $L_F=C_G L_\pi/c_0$.

For a finite $\delta$-net $\cB_\delta$ of $\cB$,
\[
\sup_{\beta\in\cB}
|(\mathbb Q_{n,T_j+1}-Q_{T_j+1})F_\beta|
\leq
\max_{\beta\in\cB_\delta}
|(\mathbb Q_{n,T_j+1}-Q_{T_j+1})F_\beta|
+2L_F\delta.
\]
By (i), the maximum converges to zero in probability; letting $\delta\downarrow0$ establishes uniform convergence.

It remains to identify the limiting matrices. Under $Q_{T_j+1}$, the marginal distribution of $S_t$ is $\eta_t^\circ$, and conditional on $S_t=s$, the treatment $A_t$ has distribution $\mu_t^\circ(\cdot;s)$. Consequently,
\begin{align*}
&\E_{Q_{T_j+1}}\left[
\frac{\pi_\beta(A_t;S_t)}{\mu_t^\circ(A_t;S_t)}
\Phi(S_t)\{\Phi(S_t)-\gamma\Phi(S_{t+1})\}^\top
\right]\\
&\qquad=
\int \limits_{\cS}\sum \limits_{a\in\cA}
\pi_\beta(a;s)\Phi(s)
\{\Phi(s)-\gamma(P_a\Phi)(s)\}^\top
\,d\eta_t^\circ(s),
\end{align*}
where the behavior probabilities cancel upon conditioning on $S_t$. Similarly,
\[
\E_{Q_{T_j+1}}\left[
\frac{\pi_\beta(A_t;S_t)}{\mu_t^\circ(A_t;S_t)}
U_t\Phi(S_t)
\right]
=
\int \limits_{\cS}\sum \limits_{a\in\cA}
\pi_\beta(a;s)r_a(s)\Phi(s)
\,d\eta_t^\circ(s).
\]
Averaging these identities over $t=1,\ldots,T_j$ gives \eqref{eq:population-W} and \eqref{eq:population-b}, respectively. Together with the uniform convergence established above, this proves \eqref{eq:matrix-consistency} for batch $j$.

\item By Assumption~\ref{ass:identifiability} and (ii), with probability tending to one,
\[
\sup_{\beta\in\cB}
\|\hat W_{\mathrm{cum},j}(\pi_\beta)^{-1}\|
\leq 2c_{1,j}^{-1}.
\]
On this event, the penalized criterion in \eqref{eq:theta-est} is continuous, strictly convex, and coercive on the closed convex set $\Theta$. Indeed, uniform invertibility and boundedness of $\hat b_{\mathrm{cum},j}$ imply that its quadratic part tends to infinity uniformly in $\beta$ as $\|\theta\|\to\infty$. Since $J$ is nonnegative and convex, the criterion has a unique minimizer for every $\beta$. Uniform coercivity, continuity of the criterion in $(\beta,\theta)$, and uniqueness also imply continuity of $\beta\mapsto\hat\theta_{\pi_\beta,n}^j$.

Write $\theta_\beta=\theta_0^{\pi_\beta}$,
$\hat\theta_\beta=\hat\theta_{\pi_\beta,n}^j$, and
$\Lambda_\beta(\theta)=\Lambda_{n,j}(\pi_\beta,\theta)$.
Lemma~\ref{lem:population} and (ii) give
\[
\sup_{\beta\in\cB}
\|\Lambda_\beta(\theta_\beta)\|
=o_{\mathbb{P}}(1).
\]
The set $\{\theta_\beta:\beta\in\cB\}$ is compact and contained in $\Theta$ by Assumption~\ref{ass:specification}; hence
$\sup_\beta J(\theta_\beta)<\infty$.
By the minimizing property of $\hat\theta_\beta$,
\begin{align*}
\lambda_{\min}(\Omega)
\|\Lambda_\beta(\hat\theta_\beta)\|^2
&\leq
\Lambda_\beta(\hat\theta_\beta)^\top
\Omega\Lambda_\beta(\hat\theta_\beta)
+\lambda_{n,j}J(\hat\theta_\beta)\\
&\leq
\lambda_{\max}(\Omega)
\|\Lambda_\beta(\theta_\beta)\|^2
+\lambda_{n,j}J(\theta_\beta).
\end{align*}
Since $\lambda_{n,j}\to0$,
\[
\sup_{\beta\in\cB}
\|\Lambda_\beta(\hat\theta_\beta)\|
=o_{\mathbb{P}}(1).
\]
Using the linearity of $\Lambda_\beta$,
\[
\hat W_{\mathrm{cum},j}(\pi_\beta)
(\hat\theta_\beta-\theta_\beta)
=
\Lambda_\beta(\theta_\beta)
-\Lambda_\beta(\hat\theta_\beta).
\]
The uniform inverse bound therefore yields
\[
\sup_{\beta\in\cB}
\|\hat\theta_\beta-\theta_\beta\|
=o_{\mathbb{P}}(1).
\]

It follows that
\[
\sup_{\beta\in\cB}
|\hat V_{n,R,j}(\pi_\beta)-M(\beta)|
=o_{\mathbb{P}}(1).
\]
By Lemma~\ref{lem:population}, $M$ is continuous on the compact set $\cB$. Assumption~\ref{ass:unique} implies that, for every $\epsilon>0$ such that the constraint set is nonempty,
\[
\inf_{\beta\in\cB:\,
\|\beta-\beta^*\|\geq\epsilon}
\{M(\beta^*)-M(\beta)\}>0.
\]
The argmax theorem \citep[Theorem~5.7]{vdV1998} now gives
$\hat\beta_n^j\xrightarrow{\mathbb{P}}\beta^*$.

This establishes the induction hypothesis for batch $j+1$ and completes the proof.
\end{enumerate}

\end{proof}

\phantomsection\label{prf:lem4.2}
\begin{proof}[Proof of Lemma~\ref{lem:equicontinuity}]
Fix a coordinate $r\in\{1,\ldots,q\}$. By Lemma~\ref{lem:population}, the map $\beta\mapsto\theta_0^{\pi_\beta}$ is bounded and Lipschitz. The logistic probabilities are Lipschitz uniformly in $(a,s)$, and Assumptions~\ref{ass:state}--\ref{ass:identifiability} give bounded rewards, features, and importance weights. Consequently, there is $L<\infty$, independent of $n,t,i$, and the realized data, such that
\[
|\xi_t^i(\beta)_r-\xi_t^i(\beta')_r|\le L\|\beta-\beta'\|.
\]
By \eqref{eq:martingale-centering}, each difference is a martingale difference. Conditional Hoeffding bounds (\citealp{Hoeffding1963}; \citealp{Azuma1967}), applied successively to the $N_k$ normalized increments, give
\begin{equation}\label{eq:increment-tail}
\P\left\{|\mathbb G_{n,k}(\beta)_r-\mathbb G_{n,k}(\beta')_r|>x\right\}
\le 2\exp\left\{-\frac{x^2}{2L^2\|\beta-\beta'\|^2}\right\},\qquad x>0.
\end{equation}
Thus the process $\mathbb G_{n,k}(\cdot)_r$ is sub-Gaussian for the semimetric $d(\beta,\beta')=L\|\beta-\beta'\|$. Its sample paths are continuous and therefore separable, since $\cB$ is compact. Corollary~2.2.8 of \citet{vdVW1996}, applied to \eqref{eq:increment-tail}, yields
\[
\E\sup_{d(\beta,\beta')\le L\delta}
|\mathbb G_{n,k}(\beta)_r-\mathbb G_{n,k}(\beta')_r|
\le C\int \limits_0^{L\delta}\sqrt{\log N(\epsilon,\cB,d)}\,d\epsilon.
\]

Writing $d_\beta=p(K-1)$, compactness of $\cB$ gives
$N(\epsilon,\cB,d)\leq(C_0/\epsilon)^{d_\beta}$ for sufficiently small $\epsilon$. Hence, for each coordinate $r$,
\[
\begin{aligned}
\E\sup_{\substack{\beta,\beta'\in\cB\\
\|\beta-\beta'\|\leq\delta}}
|\mathbb G_{n,k}(\beta)_r-\mathbb G_{n,k}(\beta')_r| \leq
C\int \limits_0^{L\delta}
\sqrt{d_\beta\log(C_0/\epsilon)}\,d\epsilon
\leq C_1\delta\sqrt{\log(C_2/\delta)},
\end{aligned}
\]
where the constants are independent of $n$. The last inequality follows by the change of variables $\epsilon=L\delta u$ and the integrability of $\sqrt{\log(1/u)}$ on $(0,1)$. Since $q$ is fixed and $\|x\|\leq\sum \limits_{r=1}^q|x_r|$, Markov's inequality yields, for every $\varepsilon>0$,
\[
\begin{aligned}
\sup_{n\geq1}\mathbb P\left(
\sup_{\substack{\beta,\beta'\in\cB\\
\|\beta-\beta'\|\leq\delta}}
\|\mathbb G_{n,k}(\beta)-\mathbb G_{n,k}(\beta')\|
>\varepsilon\right) \leq
\frac{qC_1}{\varepsilon}
\delta\sqrt{\log(C_2/\delta)}
\longrightarrow0
\qquad\text{as }\delta\downarrow0,
\end{aligned}
\]
which proves the assertion.

\end{proof}

\phantomsection\label{prf:thm4.3}
\begin{proof}[Proof of Theorem~\ref{thm:functional-clt}]
Fix $k$. 

For any strictly positive behavior policy $\mu$, let $g_\mu(\beta,\beta';s)$ denote the expression defining $g_t(\beta,\beta';s)$ above, with $\mu(a;s)$ replacing $\mu_t^\circ(a;s)$ in the denominator. In particular, $g_{\mu_t^\circ}=g_t$.

By \eqref{eq:martingale-centering}, each finite vector formed from $\mathbb G_{n,k}(\beta_1),\ldots,\mathbb G_{n,k}(\beta_m)$ is a normalized sum of bounded martingale differences. Its conditional quadratic variation has $(\beta,\beta')$ block
\begin{align*}
&\frac1{N_k}\sum \limits_{t=1}^{T_k}\sum \limits_{i=1}^n
\E\{\xi_t^i(\beta)\xi_t^i(\beta')^\top\mid\cF_{t,i-1}\}\\
&\qquad=\frac1{T_k}\sum \limits_{t=1}^{T_k}\frac1n\sum \limits_{i=1}^n g_{\mu_{n,t}}(\beta,\beta';S_t^i).
\end{align*} To establish convergence to $\Gamma_k(\beta,\beta')$, fix $t$ and write $\eta_{n,t}=n^{-1}\sum \limits_{i=1}^n\delta_{S_t^i}$. Uniform positivity and convergence of the behavior policies imply
\[
\left\|\eta_{n,t}\{g_{\mu_{n,t}}(\beta,\beta';\cdot)-g_{\mu_t^\circ}(\beta,\beta';\cdot)\}\right\|=o_{\mathbb{P}}(1).
\]
Theorem~\ref{thm:consistency} gives $\eta_{n,t}\Rightarrow\eta_t^\circ$ in probability. Since $g_{\mu_t^\circ}(\beta,\beta';\cdot)$ is bounded and continuous, this implies
\[
\eta_{n,t}g_{\mu_t^\circ}(\beta,\beta';\cdot)
\xrightarrow{\mathbb{P}}\int \limits_{\cS}g_t(\beta,\beta';s)\,d\eta_t^\circ(s).
\]

Combining the preceding two convergences, for each fixed $t\leq T_k$,
\[
\eta_{n,t}g_{\mu_{n,t}}(\beta,\beta';\cdot)
\xrightarrow{\mathbb P}
\int \limits_{\cS}g_t(\beta,\beta';s)\,d\eta_t^\circ(s).
\]
Consequently,
\begin{align*}
&\left\|
\frac{1}{T_k}\sum \limits_{t=1}^{T_k}
\eta_{n,t}g_{\mu_{n,t}}(\beta,\beta';\cdot)
-\Gamma_k(\beta,\beta')
\right\|\\
&\quad\leq
\frac{1}{T_k}\sum \limits_{t=1}^{T_k}
\left\|
\eta_{n,t}g_{\mu_{n,t}}(\beta,\beta';\cdot)
-\int \limits_{\cS}g_t(\beta,\beta';s)\,d\eta_t^\circ(s)
\right\|
\xrightarrow{\mathbb P}0,
\end{align*}
since $T_k$ is fixed and each summand converges to zero in probability. This establishes convergence of the predictable quadratic variation to $\Gamma_k(\beta,\beta')$.

The continuity of $g_t(\beta,\beta';\cdot)$ follows from the same frozen-integrand argument used to establish continuity of $K_{\mu_t^\circ}f$ in the proof of Theorem~\ref{thm:consistency}, together with the continuity and uniform positivity of the behavior policies.

 Since the martingale increments are uniformly bounded and $N_k\to\infty$, the conditional Lindeberg condition holds. The multivariate martingale central limit theorem (see, e.g., \citealp[Chapter~3]{HallHeyde1980}) yields joint Gaussian convergence at each finite set of policy parameters, with covariance blocks $\Gamma_k$.

Lemma~\ref{lem:equicontinuity} provides asymptotic stochastic equicontinuity in the Euclidean parameter metric. The finite-dimensional convergence and compactness of $\cB$ therefore give
\[
\mathbb G_{n,k}\rightsquigarrow\mathbb G_k\quad\text{in }\ell^\infty(\cB)^q,
\]
where $\mathbb G_k$ is a tight mean-zero Gaussian process with covariance $\Gamma_k$. In particular, $\|\mathbb G_{n,k}\|_\infty=O_{\mathbb{P}}(1)$.

By Theorem~\ref{thm:consistency}, the estimated coefficients converge uniformly to the population coefficients. Their population image is a compact subset of $\operatorname{int}(\Theta)$; hence the minimizers are interior with probability tending to one. The first-order condition for \eqref{eq:theta-est} is then
\[
2\hat W_{\mathrm{cum},k}(\pi)^\top\Omega
\Lambda_{n,k}(\pi,\hat\theta_{\pi,n}^k)
=\lambda_{n,k}\nabla J(\hat\theta_{\pi,n}^k).
\]
Since $\Lambda_{n,k}(\pi,\theta)$ is affine in $\theta$, the exact rearrangement is
\begin{align*}
\sqrt{N_k}(\hat\theta_{\pi,n}^k-\theta_0^\pi)
&=\{\hat W_{\mathrm{cum},k}(\pi)^\top\Omega\hat W_{\mathrm{cum},k}(\pi)\}^{-1}
\hat W_{\mathrm{cum},k}(\pi)^\top\Omega
\sqrt{N_k}\Lambda_{n,k}(\pi,\theta_0^\pi)\\
&\quad-\frac{\sqrt{N_k}\lambda_{n,k}}{2}
\{\hat W_{\mathrm{cum},k}(\pi)^\top\Omega\hat W_{\mathrm{cum},k}(\pi)\}^{-1}\nabla J(\hat\theta_{\pi,n}^k).
\end{align*}
Uniform consistency and the local boundedness of $\nabla J$ imply $\sup_{\pi\in\Pi}\|\nabla J(\hat\theta_{\pi,n}^k)\|=O_{\mathbb{P}}(1)$. Assumptions~\ref{ass:identifiability} and \ref{ass:penalty} make the second term $o_{\mathbb{P}}(1)$ uniformly. Since the matrices are square and invertible with probability tending to one,
\[
\{\hat W^\top\Omega\hat W\}^{-1}\hat W^\top\Omega=\hat W^{-1}.
\]
Using the uniform matrix convergence in \eqref{eq:matrix-consistency} and $\|\mathbb G_{n,k}\|_\infty=O_{\mathbb{P}}(1)$, we obtain the uniform expansion
\begin{equation}\label{eq:linearization}
\sup_{\beta\in\cB}\left\|
\sqrt{N_k}(\hat\theta_{\pi_\beta,n}^k-\theta_\beta)
-W_{\mathrm{cum},k}(\pi_\beta)^{-1}\mathbb G_{n,k}(\beta)
\right\|=o_{\mathbb{P}}(1).
\end{equation}
The functional continuous mapping theorem and \eqref{eq:score-covariance} now give \eqref{eq:functional-clt} with
$\mathbb Z_k(\beta)=W_{\mathrm{cum},k}(\pi_\beta)^{-1}\mathbb G_k(\beta)$. Multiplication by $\nu^\top$ proves the value-process assertion.

\end{proof}

\phantomsection\label{prf:cor4.4}
\begin{proof}[Proof of Corollary~\ref{cor:oracle}]
Under the oracle experiment, $\mu_{n,t}=\mu_t^\circ$ for every $n$ and $t$, and the patient histories are i.i.d.\ with law $Q_{T_k+1}$. The uniform matrix laws, martingale central limit theorem, and stochastic equicontinuity arguments in the proof of Theorem~\ref{thm:functional-clt} therefore apply without the behavior-policy approximation error. The limiting population matrices and score covariance kernel are unchanged, so the uniform linearization \eqref{eq:linearization} gives the same Gaussian limit $\mathbb Z_k$. Multiplication by $\nu^\top$ proves the value-process assertion.
\end{proof}

\phantomsection\label{prf:thm4.5}
\begin{proof}[Proof of Theorem~\ref{thm:policy-value}]

By Theorem~\ref{thm:consistency}, $\hat{\beta}_n^k \to \beta^*$ in probability. Set
\[
Z_{n,k}(\beta)=\sqrt{N_k}\{\hat V_{n,R,k}(\pi_\beta)-M(\beta)\}.
\]
Theorem~\ref{thm:functional-clt} gives weak convergence of $Z_{n,k}$ in $\ell^\infty(\cB)$ to a tight Gaussian process with continuous sample paths. Its asymptotic equicontinuity and $\hat\beta_n^k\to\beta^*$ imply
\[
Z_{n,k}(\hat\beta_n^k)-Z_{n,k}(\beta^*)=o_{\mathbb{P}}(1).
\]
Therefore $Z_{n,k}(\hat\beta_n^k)$ converges in distribution to $\nu^\top\mathbb Z_k(\beta^*)$, proving \eqref{eq:optimal-value-clt}.

For the alternative centering, the definition of $\hat\beta_n^k$ and optimality of $\beta^*$ give
\[
0\leq M(\beta^*)-M(\hat\beta_n^k)
\leq \hat V_{n,R,k}(\pi_{\hat\beta_n^k})-M(\hat\beta_n^k)
-\{\hat V_{n,R,k}(\pi_{\beta^*})-M(\beta^*)\}.
\]
After multiplication by $\sqrt{N_k}$, the upper bound is
$Z_{n,k}(\hat\beta_n^k)-Z_{n,k}(\beta^*)=o_{\mathbb{P}}(1)$. Hence
$\sqrt{N_k}\{M(\hat\beta_n^k)-M(\beta^*)\}=o_{\mathbb{P}}(1)$, and Slutsky's theorem yields the second centering.
\end{proof}

\phantomsection\label{prf:prop4.6}
\begin{proof}[Proof of Proposition~\ref{prop:variance}]
Fix $k$. For a complete history $h=(s_1,a_1,\ldots,s_{T_k+1})\in\cH_{T_k+1}$, let
\[
m_{k,\beta}(h)=\frac1{\sqrt{T_k}}\sum \limits_{t=1}^{T_k}
\frac{\pi_\beta(a_t;s_t)}{\mu_t^\circ(a_t;s_t)}
\left\{u(s_{t+1},a_t,s_t)+\gamma\Phi(s_{t+1})^\top\theta_\beta-\Phi(s_t)^\top\theta_\beta\right\}\Phi(s_t).
\]
By compactness, positivity, and Lemma~\ref{lem:population}, the functions $m_{k,\beta}$ are bounded, jointly continuous in $(h,\beta)$, and uniformly Lipschitz in $\beta$. Theorem~\ref{thm:consistency} and the uniform convergence of $\mu_{n,t}$ give, for fixed $k$,
\[
\sup_{1\leq i\leq n}\sup_{\beta\in\cB}
\|\hat\psi_i^k(\beta)-m_{k,\beta}(H_{i,T_k+1})\|=o_{\mathbb{P}}(1).
\]
The left side is bounded by a deterministic constant times the sum of the uniform errors in $\hat\theta_{\pi_\beta,n}^k$ and the finitely many behavior policies, so the bound is valid despite their dependence on the full data. Thus, uniformly in $\beta,\beta'$,
\[
\hat\Gamma_k(\beta,\beta')
-\frac1n\sum \limits_{i=1}^nm_{k,\beta}(H_{i,T_k+1})m_{k,\beta'}(H_{i,T_k+1})^\top=o_{\mathbb{P}}(1).
\]
The empirical history measures converge to $Q_{T_k+1}$ by Theorem~\ref{thm:consistency}. Uniform continuity over the compact parameter set and a finite-net argument therefore yield
\[
\sup_{\beta,\beta'}\left\|
\frac1n\sum \limits_{i=1}^nm_{k,\beta}(H_{i,T_k+1})m_{k,\beta'}(H_{i,T_k+1})^\top
-\int m_{k,\beta}(h)m_{k,\beta'}(h)^\top\,dQ_{T_k+1}(h)
\right\|=o_{\mathbb{P}}(1).
\]

Let $H^\circ_{T_k+1}$ denote a generic oracle trajectory, generated from the initial distribution $\eta_1^\circ$, the behavior policies $\{\mu_t^\circ\}$, and the transition kernels $\{P_a\}$, and let $\mathbb{E}^\circ$ denote expectation under its law, so that $\mathbb{E}^\circ[f(H^\circ_t)] = \int f(h)\,dQ_t(h)$ for every integrable $f$. Let $\xi_t^\circ(\beta)$ denote the time-$t$ summand in $\sqrt{T_k}\,m_{k,\beta}(H^\circ_{T_k+1})$, and define $\mathcal{F}^\circ_{t,0} = \sigma(H^\circ_t)$. The conditional Bellman equation gives
\[
\E^\circ[\xi_t^\circ(\beta)\mid\cF_{t,0}^\circ]=0.
\]
In particular, for $t<s$, $\xi_t^\circ(\beta)$ is measurable with respect to the information available before treatment at time $s$, and hence
\[
\E^\circ\!\left[
\xi_t^\circ(\beta)\xi_s^\circ(\beta')^\top
\right]
=
\E^\circ\!\left[
\xi_t^\circ(\beta)
\E^\circ\left[\xi_s^\circ(\beta')^\top\mid\cF_{s,0}^\circ\right]
\right]
=0.
\]
Similarly, conditioning on $S_t$ and summing over the treatment assignments gives
\[
\E^\circ\!\left[
\xi_t^\circ(\beta)\xi_t^\circ(\beta')^\top
\right]
=
\int \limits_{\cS}g_t(\beta,\beta';s)\,d\eta_t^\circ(s).
\]
Therefore, expanding the product of the sums defining $m_{k,\beta}$ and using the vanishing cross-time terms,
\begin{align*}
&\int m_{k,\beta}(h)m_{k,\beta'}(h)^\top
\,dQ_{T_k+1}(h)\\
&\qquad=
\frac{1}{T_k}\sum \limits_{t=1}^{T_k}
\int \limits_{\cS}g_t(\beta,\beta';s)\,d\eta_t^\circ(s)
=\Gamma_k(\beta,\beta').
\end{align*}

We have proved $\sup_{\beta,\beta'}\|\hat\Gamma_k(\beta,\beta')-\Gamma_k(\beta,\beta')\|=o_{\mathbb{P}}(1)$. Uniform matrix consistency, Assumption~\ref{ass:identifiability}, and the inverse-matrix identity yield uniform convergence of the corresponding covariances. Continuity in $\beta$ and $\hat\beta_n^k\to\beta^*$ prove the scalar variance claim. Studentization follows from Slutsky's theorem when $\sigma_{V,k}^2>0$.

\end{proof}

\section{Proofs for Section~\ref{sec:fixed-n}}\label{app:fixed-n}

\phantomsection\label{prf:lem5.1}
\begin{proof}[Proof of Lemma~\ref{lem:longrun-slln}]
Fix $\beta\in\cB$ and a coordinate $r\in\{1,\ldots,d\}$. By the conditional mean-zero property, $\{X_t(\beta)_r\}_{t\geq1}$ is a martingale-difference sequence with increments bounded by $C$. The Azuma--Hoeffding inequality gives, for every $\epsilon>0$,
\[
\mathbb P\left(
\left|\frac1T\sum \limits_{t=1}^TX_t(\beta)_r\right|>\epsilon
\right)
\leq 2\exp\left(-\frac{T\epsilon^2}{2C^2}\right).
\]
The right-hand side is summable in $T$. Hence, by the Borel--Cantelli lemma,
\[
\frac1T\sum \limits_{t=1}^TX_t(\beta)_r
\xrightarrow{\mathrm{a.s.}}0.
\]
Since $d$ is fixed, the same conclusion holds for the vector $X_t(\beta)$.

To obtain uniform convergence, fix $\delta>0$ and let $\cB_\delta$ be a finite $\delta$-net of $\cB$. By the Lipschitz bound,
\[
\sup_{\beta\in\cB}
\left\|\frac1T\sum \limits_{t=1}^TX_t(\beta)\right\|
\leq
\max_{\beta\in\cB_\delta}
\left\|\frac1T\sum \limits_{t=1}^TX_t(\beta)\right\|
+L\delta.
\]
The maximum converges to zero almost surely, since $\cB_\delta$ is finite. Consequently,
\[
\limsup_{T\to\infty}
\sup_{\beta\in\cB}
\left\|\frac1T\sum \limits_{t=1}^TX_t(\beta)\right\|
\leq L\delta
\qquad\text{almost surely}.
\]
Applying this argument to a countable sequence $\delta\downarrow0$ proves the assertion.

\end{proof}

\phantomsection\label{prf:thm5.2}
\begin{proof}[Proof of Theorem~\ref{thm:longrun-consistency}]
Fix $n\geq1$ throughout and write $\theta_\beta=\theta_0^{\pi_\beta}$.

\emph{Proof of (i).}
For each $\beta\in\cB$, define
\[
A_t(\beta)=\frac1n\sum \limits_{i=1}^n
\frac{\pi_\beta(A_t^i;S_t^i)}
{\mu_{n,t}(A_t^i;S_t^i)}
\Phi(S_t^i)
\{\Phi(S_t^i)-\gamma\Phi(S_{t+1}^i)\}^\top.
\]
Since the states and behavior policy at time $t$ are $\cF_{t,0}$-measurable, Assumption~\ref{ass:randomization} gives
\[
\begin{aligned}
\E[A_t(\beta)\mid\cF_{t,0}]
&=\frac1n\sum \limits_{i=1}^n\sum \limits_{a\in\cA}
\pi_\beta(a;S_t^i)\Phi(S_t^i)
\{\Phi(S_t^i)-\gamma(P_a\Phi)(S_t^i)\}^\top.
\end{aligned}
\]
Thus, writing
$X_t(\beta)=A_t(\beta)-\E[A_t(\beta)\mid\cF_{t,0}]$,
we have
\[
\hat W_{\mathrm{cum},k}(\pi_\beta)
-\overline W_{n,k}(\pi_\beta)
=\frac1{T_k}\sum \limits_{t=1}^{T_k}X_t(\beta).
\]
Each coordinate of $X_t(\beta)$ is a martingale difference. Moreover, boundedness of $\Phi$, uniform positivity of the behavior policies, and the Lipschitz property of $\pi_\beta$ imply that $X_t(\beta)$ is uniformly bounded and Lipschitz in $\beta$, with constants independent of $t$. Applying Lemma~\ref{lem:longrun-slln} to the vectorized matrices, and noting that $T_k\to\infty$, yields
\[
\sup_{\beta\in\cB}
\|\hat W_{\mathrm{cum},k}(\pi_\beta)
-\overline W_{n,k}(\pi_\beta)\|
\xrightarrow{\mathrm{a.s.}}0,
\]
which proves (i).

\emph{Proof of (ii).}
We first establish uniform convergence of the estimating function at the population coefficients. For $\beta\in\cB$, let
\[
D_t(\beta)=\frac1n\sum \limits_{i=1}^n
\frac{\pi_\beta(A_t^i;S_t^i)}
{\mu_{n,t}(A_t^i;S_t^i)}
e_t^i(\theta_\beta)\Phi(S_t^i).
\]
By the conditional treatment-assignment mechanism,
\[
\begin{aligned}
\E[D_t(\beta)\mid\cF_{t,0}]
&=\frac1n\sum \limits_{i=1}^n\Phi(S_t^i)
\sum \limits_{a\in\cA}\pi_\beta(a;S_t^i)\\
&\quad\times\left\{
r_a(S_t^i)
+\gamma(P_a\Phi)(S_t^i)^\top\theta_\beta
-\Phi(S_t^i)^\top\theta_\beta
\right\}
=0,
\end{aligned}
\]
where the last equality follows from the Bellman equation under Assumption~\ref{ass:specification}. Thus $D_t(\beta)$ is a martingale difference. By Lemma~\ref{lem:population}, $\theta_\beta$ is bounded and Lipschitz in $\beta$. Together with the boundedness and positivity conditions, this implies that $D_t(\beta)$ satisfies the hypotheses of Lemma~\ref{lem:longrun-slln}. Consequently,
\[
\sup_{\beta\in\cB}
\left\|\frac1T\sum \limits_{t=1}^TD_t(\beta)\right\|
\xrightarrow{\mathrm{a.s.}}0.
\]
Since $N_k=nT_k$, the definition of $\Lambda_{n,k}$ gives
\[
\Lambda_{n,k}(\pi_\beta,\theta_\beta)
=\frac1{T_k}\sum \limits_{t=1}^{T_k}D_t(\beta).
\]
The preceding convergence therefore holds along $\{T_k\}$. In particular,
\begin{equation}\label{eq:longrun-score}
R_k:=
\sup_{\beta\in\cB}
\|\Lambda_{n,k}(\pi_\beta,\theta_\beta)\|
\xrightarrow{\mathrm{a.s.}}0.
\end{equation}

By Assumption~\ref{ass:persistent} and part (i),
\[
\begin{aligned}
\liminf_{k\to\infty}\inf_{\beta\in\cB}
\sigma_{\min}\{\hat W_{\mathrm{cum},k}(\pi_\beta)\} &\geq
\liminf_{k\to\infty}\inf_{\beta\in\cB}
\sigma_{\min}\{\overline W_{n,k}(\pi_\beta)\}\\
&\qquad-
\limsup_{k\to\infty}\sup_{\beta\in\cB}
\|\hat W_{\mathrm{cum},k}(\pi_\beta)
-\overline W_{n,k}(\pi_\beta)\|
>0
\end{aligned}
\]
almost surely, where the inequality follows from
$\sigma_{\min}(A)\geq\sigma_{\min}(B)-\|A-B\|$.
Consequently, the empirical matrices are eventually uniformly invertible, with
\begin{equation}\label{eq:longrun-inverse}
\sup_{\beta\in\cB}
\|\hat W_{\mathrm{cum},k}(\pi_\beta)^{-1}\|
=O_{\mathrm{a.s.}}(1).
\end{equation}
Here the bound may depend on the sample path, but is finite and independent of $k$ for all sufficiently large $k$.

We next use the minimizing property of $\hat\theta_{\pi_\beta,n}^k$. Since $\theta_\beta\in\Theta$ and $J\geq0$,
\[
\begin{aligned}
&\Lambda_{n,k}(\pi_\beta,\hat\theta_{\pi_\beta,n}^k)^\top
\Omega\Lambda_{n,k}(\pi_\beta,\hat\theta_{\pi_\beta,n}^k)\\
&\qquad\leq
\Lambda_{n,k}(\pi_\beta,\theta_\beta)^\top
\Omega\Lambda_{n,k}(\pi_\beta,\theta_\beta)
+\lambda_{n,k}J(\theta_\beta).
\end{aligned}
\]
By Lemma~\ref{lem:population}, the set
$\{\theta_\beta:\beta\in\cB\}$ is compact. Continuity of $J$ therefore gives
$C_J:=\sup_{\beta\in\cB}J(\theta_\beta)<\infty$.
Using the positive definiteness of $\Omega$ and taking the supremum over $\beta$, we obtain
\[
\begin{aligned}
\sup_{\beta\in\cB}
\|\Lambda_{n,k}(\pi_\beta,\hat\theta_{\pi_\beta,n}^k)\|^2 \leq
\frac{\lambda_{\max}(\Omega) R_k^2 + \lambda_{n,k}C_J}{\lambda_{\min}(\Omega)}
\xrightarrow{\mathrm{a.s.}}0,
\end{aligned}
\]
by \eqref{eq:longrun-score} and $\lambda_{n,k}\to0$.

Finally, the affine representation of the estimating function gives
\[
\begin{aligned}
\hat W_{\mathrm{cum},k}(\pi_\beta)
(\hat\theta_{\pi_\beta,n}^k-\theta_\beta) =
\Lambda_{n,k}(\pi_\beta,\theta_\beta)
-\Lambda_{n,k}(\pi_\beta,\hat\theta_{\pi_\beta,n}^k).
\end{aligned}
\]
Thus, for all sufficiently large $k$,
\[
\begin{aligned}
\sup_{\beta\in\cB}
\|\hat\theta_{\pi_\beta,n}^k-\theta_\beta\| \leq
\sup_{\beta\in\cB}
\|\hat W_{\mathrm{cum},k}(\pi_\beta)^{-1}\|
\left\{
R_k+
\sup_{\beta\in\cB}
\|\Lambda_{n,k}(\pi_\beta,\hat\theta_{\pi_\beta,n}^k)\|
\right\}
\xrightarrow{\mathrm{a.s.}}0,
\end{aligned}
\]
by \eqref{eq:longrun-inverse} and the two preceding convergence results. This proves (ii).

\emph{Proof of (iii).}
By part (ii) and the definitions of $\hat V_{n,R,k}$ and $M$,
\[
\begin{aligned}
\sup_{\beta\in\cB}
|\hat V_{n,R,k}(\pi_\beta)-M(\beta)|
&\leq
\|\nu\|
\sup_{\beta\in\cB}
\|\hat\theta_{\pi_\beta,n}^k-\theta_\beta\|\\
&\xrightarrow{\mathrm{a.s.}}0.
\end{aligned}
\]
Since $\hat\beta_n^k$ maximizes $\hat V_{n,R,k}$ over $\cB$ and $\beta^*$ maximizes $M$, we have
\[
\begin{aligned}
0\leq M(\beta^*)-M(\hat\beta_n^k)
\leq
2\sup_{\beta\in\cB}
|\hat V_{n,R,k}(\pi_\beta)-M(\beta)|
\xrightarrow{\mathrm{a.s.}}0.
\end{aligned}
\]
For any $\epsilon>0$, compactness of $\cB$, continuity of $M$, and uniqueness of $\beta^*$ imply
\[
\sup_{\substack{\beta\in\cB\\
\|\beta-\beta^*\|\geq\epsilon}}M(\beta)
<M(\beta^*)
\]
whenever the constraint set is nonempty. The preceding convergence therefore forces
$\|\hat\beta_n^k-\beta^*\|<\epsilon$
eventually almost surely. Since $\epsilon$ is arbitrary,
$\hat\beta_n^k\xrightarrow{\mathrm{a.s.}}\beta^*$,
establishing (iii).

\end{proof}

\bibliographystyle{plainnat}
\bibliography{references}

\end{document}